\documentclass[a4paper,fleqn]{cas-sc}

\usepackage{bbm}
\usepackage{amsmath}
\usepackage{soul}  
\usepackage{pifont}
\usepackage{algorithm}
\usepackage{algorithmic}
\newcommand{\cmark}{\ding{51}} 
\newcommand{\xmark}{\ding{55}} 
\usepackage[numbers,sort&compress]{natbib}
\def\tsc#1{\csdef{#1}{\textsc{\lowercase{#1}}\xspace}}
\tsc{WGM}
\tsc{QE}
\tsc{EP}
\tsc{PMS}
\tsc{BEC}
\tsc{DE}

\begin{document}
\let\WriteBookmarks\relax
\def\floatpagepagefraction{1}
\def\textpagefraction{.001}

\shorttitle{Cognitive Disentanglement for Referring Multi-Object Tracking}

\shortauthors{Liang et~al.}

\title [mode = title]{Cognitive Disentanglement for Referring Multi-Object Tracking}                      

%
\author[1,3]{Shaofeng Liang}[
style=chinese,
orcid=0009-0005-8134-2188
]

\fnmark[1]
\ead{s23070053@s.upc.edu.cn}
\credit{Conceptualization, Methodology, Experiments, Paper Writing}

\author[2,9]{Runwei Guan}[
style=chinese,
orcid=0000-0003-4013-2107
]
\cormark[1]
\fnmark[1]
\ead{runwei.guan@liverpool.ac.uk}
\credit{Conceptualization, Methodology, Paper Writing}
\author[1,3]{Wangwang Lian}[
style=chinese,
orcid=0009-0002-4347-2617
]
\ead{2207020110@s.upc.edu.cn}
\credit{Paper Review and Correction}

\author[4]{Daizong Liu}[
style=chinese,
orcid=0000-0001-8179-4508
]
\ead{dzliu@stu.pku.edu.cn}
\credit{Paper Review and Correction}

\author[5]{Xiaolou Sun}[
style=chinese,
orcid=0009-0007-3597-3872
]
\ead{xlsun@seu.edu.cn}
\credit{Paper Review and Correction}

\author[6]{Dongming Wu}[style=chinese,orcid=0000-0003-4938-5813]
\ead{wudongming97@gmail.com}
\credit{Paper Review and Correction}

\author%
[2,8]
{Yutao Yue}[style=chinese, orcid=0000-0003-4532-0924]
\ead{yutaoyue@hkust-gz.edu.cn}
\credit{Supervision}

\author[7]{Weiping Ding}[style=chinese,orcid=0000-0002-3180-7347]
\ead{ding.wp@ntu.edu.cn}
\credit{Paper Review and Correction}

\author[2]{Hui Xiong}[style=chinese,orcid=0000-0001-6016-6465]
\ead{xionghui@ust.hk}
\credit{Supervision}

\affiliation[1]{organization={Qingdao Institute of Software, College of Computer Science and Technology, China University of Petroleum (East China)},
	city={Qingdao},
	country={China}}

\affiliation[2]{organization={Thrust of AI, Hong Kong University of Science and Technology (GuangZhou)},
	city={Guangzhou},
	country={China}}

\affiliation[3]{organization={Shandong Key Laboratory of Intelligent Oil \& Gas Industrial Software},
	city={Qingdao},
	country={China}}

\affiliation[4]{organization={Wangxuan Institute of Computer Technology, Peking University},
	city={Beijing},
	country={China}}

\affiliation[5]{organization={School of Automation, Southeast University},
	city={Nanjing},
	country={China}}

\affiliation[6]{organization={School of Computer Science, Beijing Institute of Technology},
	city={Beijing},
	country={China}}
    
\affiliation[7]{organization={School of Artificial Intelligence and Computer Science, Nantong University},
	city={Nantong},
	country={China}}

\affiliation[8]{organization={Thrust of Intelligent Transportation, Hong Kong University of Science and Technology (GuangZhou)},
	city={Guangzhou},
	country={China}}

\affiliation[9]{organization={Department of Electrical Engineering and Electronics, University of Liverpool},
	city={Liverpool},
	country={United Kingdom}}

\cortext[cor1]{Corresponding author}

\begin{abstract}
As a significant application of multi-source information fusion in intelligent transportation perception systems, Referring Multi-Object Tracking (RMOT) involves localizing and tracking specific objects in video sequences based on language references. However, existing RMOT approaches often treat language descriptions as holistic embeddings and struggle to effectively integrate the rich semantic information contained in language expressions with visual features. This limitation is especially apparent in complex scenes requiring comprehensive understanding of both static object attributes and spatial motion information. In this paper, we propose a Cognitive Disentanglement for Referring Multi-Object Tracking (CDRMT) framework that addresses these challenges. It adapts the "what" and "where" pathways from the human visual processing system to RMOT tasks. Specifically, our framework first establishes cross-modal connections while preserving modality-specific characteristics. It then disentangles language descriptions and hierarchically injects them into object queries, refining object understanding from coarse to fine-grained semantic levels. Finally,  we reconstruct language representations based on visual features, ensuring that tracked objects faithfully reflect the referring expression.
Extensive experiments on different benchmark datasets demonstrate that CDRMT achieves substantial improvements over state-of-the-art methods, with average gains of 6.0\% in HOTA score on Refer-KITTI and 3.2\% on Refer-KITTI-V2. Our approach advances the state-of-the-art in RMOT while simultaneously providing new insights into multi-source information fusion.
\end{abstract}

\begin{keywords}
referring multi-object tracking \sep human-centric perception  \sep human-visual-inspired neural network\sep  vision–language fusion 
\end{keywords}

\maketitle

\section{Introduction}

In recent years, the rapid development of intelligent transportation perception systems has been driven by multi-source information fusion, which has become a prominent research focus in computer vision~\cite{IF1,IF2,IF3,IF4,IF6}. The integration of visual and language information~\cite{IF8,IF9} has catalyzed significant breakthroughs across various tasks, including visual question-answering~\cite{vqa}, image captioning~\cite{imagecaption}, and visual navigation~\cite{navigation}.
In this context, Referring Multi-Object Tracking (RMOT)~\cite{rmot1}, as an emerging multimodal task, is receiving widespread attention  such as human-computer interaction~\cite{IF5}, autonomous driving~\cite{IF7,IF10} and embodied intelligence~\cite{IF6} domain. 

\begin{figure}[!b]
	\centering
	\vspace{-0.6cm}
	\includegraphics[width=\linewidth]{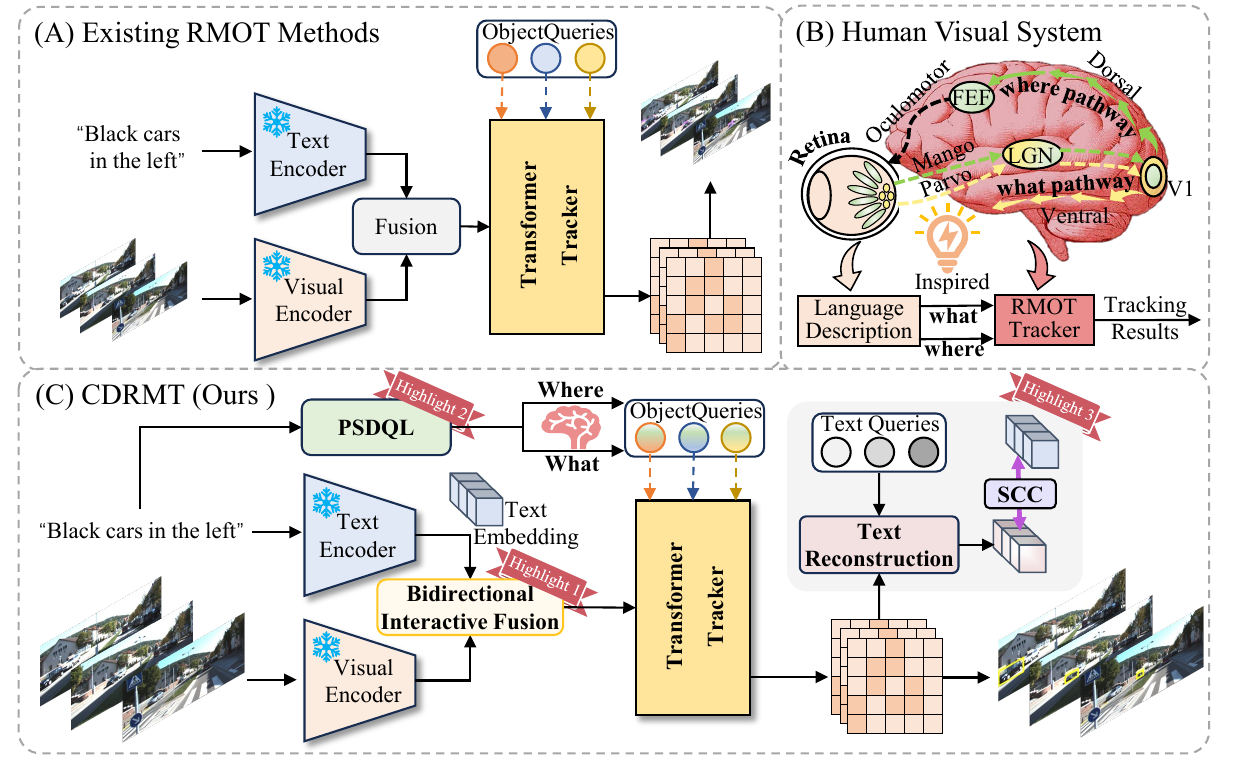}
	\vspace{-0.6cm}
	\hfill
	\caption{From existing RMOT methods to our CDRMT framework by human visual system inspiration. (A) Conventional RMOT approaches typically employ a unified architecture where visual and textual features are plainly fused before tracking. (B) The human visual system processes information through two distinct pathways: the ventral ("what") stream for object recognition and the dorsal ("where") stream for spatial processing, which serves as the biological inspiration for our method. (C) Our proposed Cognitive Disentanglement for Referring Multi-Object Tracking (CDRMT) framework explicitly decouples and separately processes natural language description information, while introducing structural consistency constraints to enhance bidirectional semantic understanding between vision and language.}
	\label{fig:motivation}
	\vspace{-0.7cm}
\end{figure}

Conventional Multi-Object Tracking (MOT) methods~\cite{MOT1,MOT2,MOT3,MOT4,motr,motrv2} focus on detecting and tracking all visible objects in video sequences, while RMOT innovatively introduces language descriptions as selective tracking instructions, enabling precise tracking of specific objects. This language-guided paradigm transcends the conventional limitations of predefined categories and manual annotations, providing users with an intuitive and natural interaction method. 
Recent research~\cite{rmot2,mls-track,rmot-tim,lamot,guan2025watervg,guan2025referring,crossviewrmot} has extensively explored various approaches to enhance model localization capabilities in RMOT task. This includes language-visual fusion strategies~\cite{rmoticassp} that aim to effectively integrate linguistic information with visual features, text-conditional query mechanisms~\cite{ros1}  that leverage language descriptions to guide tracking, and post-processing optimization~\cite{ikun} techniques to refine tracking results. 
Although these methods have achieved promising results~\cite{rmoticassp, ikun, ros1, rmot2, rmot1}, they generally treat referring expressions as holistic embeddings, lacking effective mechanisms to handle the various fine-grained semantic information within referring expressions,  as shown in  Figure~\ref{fig:motivation}-A.
This plain fusion method ignores the potential semantic diversity in language description and the temporal dynamics of visual information, leading to suboptimal tracking feature representations. Inspired by the dual-stream processing mechanism in human visual system~\cite{brain1,brain3}, we rethink the language-visual information processing mechanism in RMOT tasks.

In cognitive neuroscience~\cite{brain2,brain4}, visual processing begins in the retina through two types of retinal ganglion cells: magnocellular (green) and parvocellular (yellow) cells. As Figure \ref{fig:motivation}-B illustrates, these cells encode complementary visual information and project to separate layers in the Lateral Geniculate Nuclei (LGN), which then relay to different neurons in the primary visual cortex (V1), forming two distinct neural pathways.
The parvocellular pathway leads to the ventral stream ("what" pathway), extending from V1 to the inferior temporal cortex. This stream specializes in object recognition and static feature analysis, processing colors, shapes, and details to determine "what" we see. Meanwhile, the magnocellular pathway forms the dorsal stream ("where" pathway), extending from V1 to the posterior parietal cortex and the Frontal Eye Field (FEF). This stream focuses on spatial relationships and motion information, determining "where" objects are located.
This functional segregation and collaboration ensure efficient understanding and tracking of objects in dynamic scenes. Based on this theory, we observe that language descriptions in RMOT tasks also contain two fundamentally different but complementary types of information: static object attributes (\emph{e.g.}, "black car", "women") corresponding to the "what" pathway processing and spatial location and motion state information (\emph{e.g.}, "turning right", "in the counter direction of ours") corresponding to the "where" pathway.

Therefore, we propose Cognitive Disentanglement for Referring Multi-Object Tracking (CDRMT) framework, as illustrated in  Figure~\ref{fig:motivation}-C. The key innovation of CDRMT lies in its biologically inspired architecture that imitates human visual processing system. 
Specifically, we first propose a Bidirectional Interactive Fusion (BIF) module to handle cross-modal interactions. As the retinal ganglion cells perform initial sensing and filtering of visual signals before transmission to higher brain regions, BIF (Figure~\ref{fig:motivation}-C-Highlight 1) processes visual and language features before the main encoder, allowing each modality to selectively focus on relevant information while preserving its unique characteristics. 
Moreover, we design a Progressive Semantic-Decoupled Query Learning (PSDQL) module, as depicted in Figure~\ref{fig:motivation}-C-Highlight 2. Similar to the human visual system processes information through ventral and dorsal streams, PSDQL progressively injects decoupled semantic information into queries via two complementary pathways. This hierarchical injection enables queries to better capture both static and dynamic characteristics of referred objects.
Finally, we introduce the Structural Consistency Constraint (SCC) mechanism, which serves as an additional supervisory signal during training to ensure comprehensive understanding of object appearance and language descriptions. Analogous to the higher cortical areas in the human brain that send feedback signals to earlier processing regions to refine perception and ensure consistency between prediction and sensory input. SCC (Figure~\ref{fig:motivation}-C-Highlight 3) employs a text decoder to reconstruct original descriptions from encoded visual features and computes reconstruction loss between reconstructed and original expressions.
Extensive experiments on multiple benchmark datasets demonstrate that CDRMT achieves substantial performance improvements, with average gains of 6.0\% in HOTA score on Refer-KITTI and 3.2\% on Refer-KITTI-V2.
The main contributions of this work can be summarized as follows:
\begin{itemize}
    \item 
     We propose a Cognitive Disentanglement for Referring Multi-Object Tracking (CDRMT) framework that explicitly disentangles static object and spatial motion information in referring expressions, imitating the human visual processing system for effective referring multi-object tracking.
    \item
    We introduce a Bidirectional Interactive Fusion module that enables structured cross-modal interactions while preserving modality-specific characteristics. It allows for more effective alignment between visual features and language descriptions.
    \item
    We design a Progressive Semantic-Decoupled Query Learning module that injects decoupled semantic information into queries via two complementary pathways. The queries embed static object attributes and spatial motion information separately, leading to more effective object localization and tracking.
    \item 
    We introduce a Structural Consistency Constraint that enforces bidirectional semantic consistency between reconstructed and original expressions. 
\end{itemize}

The rest of this paper is organized as follows. Section~\ref{sec:rw} provides a comprehensive review of related work in visual object tracking. Section~\ref{sec:method} details our CDRMT methodology and architectural components. Section~\ref{sec:exp} presents experimental evaluations and comparative analyses. Section~\ref{sec:vis} offers visualization analyses. Section~\ref{sec:dis} examines current limitations and explores future research directions, with the conclusion provided in Section~\ref{sec:clu}.
\section{Related Work}
\label{sec:rw}
\subsection{Visual Object Tracking}
\subsubsection{Single-Object Tracking}
Visual object tracking has witnessed significant progress over the past decades. Early success was achieved by Siamese networks~\cite{SiamFC} and discriminative correlation filters~\cite{dcf1,HCF,LADCF}, which established fundamental tracking paradigms. FBACF~\cite{zhang1} proposed a Feature Block-Aware Correlation Filter for UAV tracking that addresses the limitations of conventional DCF-based methods. SiamRPN~\cite{SiamRPN} revolutionized the field by introducing region proposal network (RPN) from object detection into the Siamese tracking framework, while its successor SiamRPN++~\cite{SiamRPN++} further enhanced performance by incorporating an effective sampling strategy to address spatial invariance limitations. However, these template-matching methods, which rely on fixed template frames, struggle with appearance variations caused by occlusion and deformation. Online trackers~\cite{UPDT,UpdateNet,9269493} address this limitation by adaptively updating object templates under specific conditions, leading to improved robustness. The advent of transformers~\cite{SwinTransformer,Vit} brought another paradigm shift, with TransT~\cite{TransTsot} pioneering the use of transformer backbones in tracking tasks. Subsequently, OSTrack~\cite{OSTrack} achieved superior performance by leveraging MAE-pretrained ViT and candidate elimination strategies. GRM~\cite{GRM} further advanced the field by introducing adaptive token partition techniques, effectively addressing the object-background confusion problem prevalent in existing tracking algorithms. ETDMT~\cite{zhang2} developed an Efficient Template Distinction Modeling Tracker with Temporal Contexts that explicitly differentiates between real target and background areas in the template. Despite these advances in single-object tracking, real-world scenarios often involve multiple objects of interest, highlighting the limitations of SOT methods in handling complex scenes with multiple interactive objects. 

\subsubsection{Multi-Object Tracking}
Multi-Object Tracking (MOT), which aims to simultaneously detect and track multiple moving objects while maintaining consistent identities across frames. MOT has become crucial for applications ranging from video surveillance to autonomous driving and sports analysis. Two primary paradigms have emerged in MOT: Tracking-by-Detection~\cite{deepsort} and Joint-Detection-and-Tracking~\cite{fairmot}. The former separates detection and association into two stages, while the latter integrates them in an end-to-end manner. Various specialized benchmarks have driven progress in different scenarios, from pedestrian tracking~\cite{pets2009} (PETS2009) to more challenging domains like autonomous driving~\cite{Kitti,bdd100k} (KITTI, BDD100K), sports analysis~\cite{MOT4} (SportsMOT), and aerial tracking~\cite{Visdrone} (VisDrone). Recent transformer-based approaches~\cite{Detransformer,motr,motrv2,zhang3} have achieved state-of-the-art performance by better modeling long-range dependencies and complex object interactions. However, conventional MOT methods, which typically track all visible objects in the scene, may not align with practical applications where users are only interested in specific objects. 
\subsection{Referring Object Tracking}
\subsubsection{Referring Single-Object Tracking}
Referring Single-Object Tracking (RSOT) emerged as a rapidly evolving research domain that integrated natural language descriptions with visual object tracking to improve localization accuracy and robustness in video sequences. The seminal work by Li et al.~\cite{VLTrack1} established foundational principles for this field by introducing a groundbreaking multimodal tracking framework. This study systematically defined three distinct tracking architectures and extended two benchmark tracking datasets, providing comprehensive empirical validation of the effectiveness of language-driven tracking methodologies. Feng et al.~\cite{VLTrack2} proposed a dynamic visual-language modality aggregation mechanism designed to be compatible with various Siamese trackers, making it a versatile solution for different tracking scenarios. Recent advancements in cross-modal integration were evidenced by Li et al~\cite{VLTrack3}, who developed a unified local-global search framework from cross-modal retrieval perspectives, while Guo et al.~\cite{VLTrack4} utilized a Modality Mixer and asymmetrical ConvNet search. The work of Zhou et al.~\cite{VLTrack5} advanced relational modeling through a multi-source relation module that explicitly constructed visual-semantic relationships between reference descriptions and object images. Most recently, Ma and Wu et al.~\cite{VLTrack6} proposed temporal context decoupling, separating and integrating long-term and short-term contextual information within visual tracking frameworks. Despite the success of the referring perception methods mentioned above, recent critical analyses ~\cite{VLTrack7,VLTrack8} demonstrated that single-object perception paradigms fundamentally limited networks' capacity for holistic environmental understanding.

\subsubsection{Referring Multi-Object Tracking}
To address the problem of RSOT, Wu et al.~\cite{rmot1} first established a Transformer-based architecture (TransRMOT) that addressed cross-modal alignment and temporal consistency. However, existing approaches often faced scalability limitations when handling complex language queries. Nguyen et al.~\cite{Type-to-track} bridged this gap through their MENDER framework, introducing a Type-to-Track paradigm that employed third-order tensor decomposition to efficiently map language types to tracklet embeddings. He~\cite{rmoticassp} proposed a hierarchical fusion architecture featuring an enhanced early-fusion module, bidirectional cross-modal encoder, and dedicated decoder. Furthermore, Du et al.~\cite{ikun} advanced adaptive feature extraction via its insertable Knowledge Unification Network (iKUN), where textual guidance directly modulates visual feature selection through learnable projection matrices. Recent efforts have also expanded RMOT's applicability to complex real-world scenarios. Wu et al.~\cite{wu2023language} contributed the NuPrompt benchmark for autonomous driving, establishing PromptTrack as an end-to-end baseline with 3D-aware language prompts. Referring Video Object Segmentation (RVOS)~\cite{RVOS1,RVOS3,textreconstruction} has also significantly contributed to RMOT development. Wang et al.~\cite{RVOS2} introduced Local-Global Context Aware Transformer for language-guided video segmentation, enhancing cross-modal interaction with multi-scale contextual information. 
However, these methods lack effective mechanisms to disentangle and process the complex semantic information contained in referring expressions, which is crucial for accurate referring multi-object tracking. In contrast, our Cognitive Disentanglement for Referring Multi-Object Tracking (CDRMT) framework disentangles the rich semantic information in referring expressions, significantly improving tracking performance in complex scenarios.
\section{Method}
\label{sec:method}
\subsection{Overview}
\begin{figure*}[!t]
	\centering
	\includegraphics[width=\linewidth]{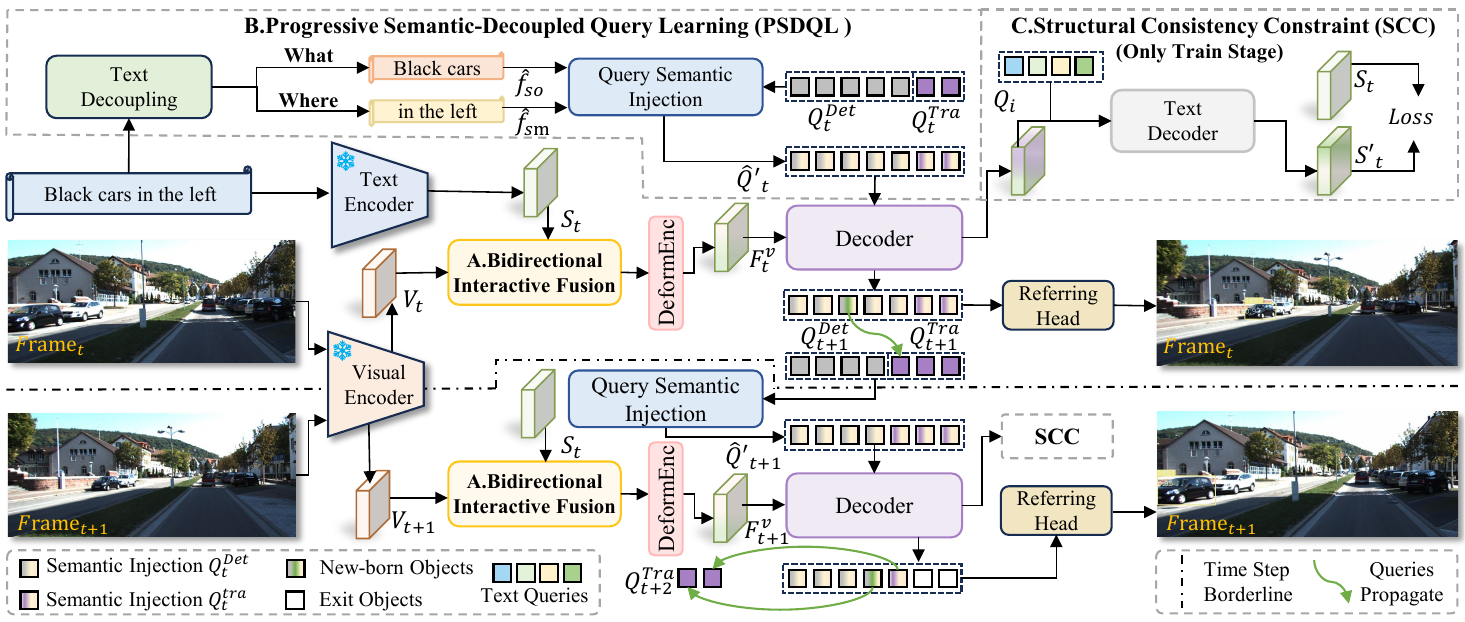}
	\vspace{-0.6cm}
	\caption{
		The architecture of our proposed Cognitive Disentanglement for Referring Multi-Object Tracking framework. It processes sequential video frames through three collaborative components: (A)The Bidirectional Interactive Fusion module first establishes cross-modal connections while preserving modality-specific characteristics. (B) The Progressive Semantic-Decoupled Query Learning (PSDQL) module, inspired by the dual-stream ("what"/"where" ) processing mechanism in human visual system, separates language into static object attributes ("what" pathway) and spatial motion information ("where" pathway) to guide object queries. (C) The Structural Consistency Constraint (SCC) mechanism is only applied during the training stage, which enforces geometric consistency between original text embeddings and their reconstructed counterparts to enhance semantic alignment between visual objects and linguistic descriptions. 
        }
	\label{fig:pipeline}
	\vspace{-0.5cm}
\end{figure*}

In this paper, we present the Cognitive Disentanglement for Referring Multi-Object Tracking (CDRMT) framework, as shown in Figure~\ref{fig:pipeline}. 
Specifically, given the $t$-th frame $\mathbf{F}_t$ of video sequence $\mathcal{V}$ and language descriptions $\mathcal{T}$, we extract visual features $\mathbf{V}_t \in \mathbb{R}^{H\times W\times C}$ and text features $\mathbf{S} \in \mathbb{R}^{L\times D}$ through pre-trained visual encoder $E_v$ and language encoder $E_l$, respectively.
To effectively align language descriptions with visual features while preserving their inherent characteristics, we propose a Bidirectional Interactive Fusion (Figure~\ref{fig:pipeline}-A) module that enables structured cross-modal interactions before the main encoder stage.
Inspired by the human visual processing mechanism, we decompose the text features into two complementary components: static object attributes $\hat{\mathbf{f}}_{so}$ corresponding to the "what" pathway, and spatial motion attributes $\hat{\mathbf{f}}_{sm}$ corresponding to the "where" pathway.
We propose a progressive semantic-decoupled query learning (Figure~\ref{fig:pipeline}-B) mechanism that the static object attributes $\hat{\mathbf{f}}_{so}$ are first injected into queries $\mathbf{Q}_\text{so}$ to localize candidate objects based on appearance features.  Subsequently, the spatial motion attributes $\hat{\mathbf{f}}_{sm}$ are incorporated into $\mathbf{Q}_\text{sm}$ to refine the query candidates, better aligning them with the referred objects in the descriptions. Finally, we introduce a text decoder $\mathcal{D}_t$ that reconstructs the original descriptions from visual features $\mathbf{F}^v_t$, and employ the Structural Consistency Constraint (SCC) mechanism to compute reconstruction loss between the generated and original expressions, as illustrated in Figure~\ref{fig:pipeline}-C. It serves as an additional constraint that enforces bidirectional semantic consistency between visual objects and language descriptions, ensuring the model's comprehensive understanding of both object appearance and linguistic attributes.
\subsection{Multimodal Feature Encoding}
\textbf{Visual Encoder}. For visual processing, we adopt the encoding paradigm of Deformable DETR~\cite{motr} with ResNet-50~\cite{resnet50} as the backbone. Given a T-frame video sequence $\{{\mathbf{F}_1, ..., \mathbf{F}_T}\}$, we extract spatio-temporal representations through the backbone encoder. Specifically, for the $t$-th frame, the feature extraction process can be formulated as:
\begin{equation}
	\begin{aligned}
		\mathbf{V}_t &= \mathcal{F}_{\text{vis}}(\mathbf{F}_t) \in \mathbb{R}^{HW \times C}, 		
	\end{aligned}
\end{equation}
where $\mathcal{F}_{\text{vis}}(\cdot)$ denotes the visual encoding function, and $C$, $H$, and $W$ represent the channel dimension, height, and width of the feature map, respectively. 

\textbf{Language Encoder}. Leveraging recent advances in language modeling, we utilize a pre-trained RoBERTa model~\cite{roberta} for textual representation. Given language descriptions $\mathcal{T}$, we first tokenize it into a token sequence $\mathbf{X} = \{x_1, ..., x_L\}$. The language encoding process can be formulated as:
\begin{equation}
		\begin{aligned}
			\mathbf{X} &= \text{Tokenize}(\mathcal{T}), \\
			\mathbf{S} &= \mathcal{F}_{\text{lan}}(\mathbf{X}) \in \mathbb{R}^{L \times D}, \\
			\mathbf{S}_g &= \text{Pool}(\mathbf{S}) \in \mathbb{R}^{1 \times D},
		\end{aligned}
            \label{eq:text}
\end{equation}
where $\mathbf{S}$ represents the token-level features with dimension $D$, $\mathcal{F}_{\text{lan}}(\cdot)$ denotes the language encoding function~\cite{roberta} and $\mathbf{S}_g$ denotes the sentence-level global representation obtained through average pooling $\text{Pool}(\cdot)$ operation.
\begin{figure*}[!t]
	\centering
	\includegraphics[width=0.8\linewidth]{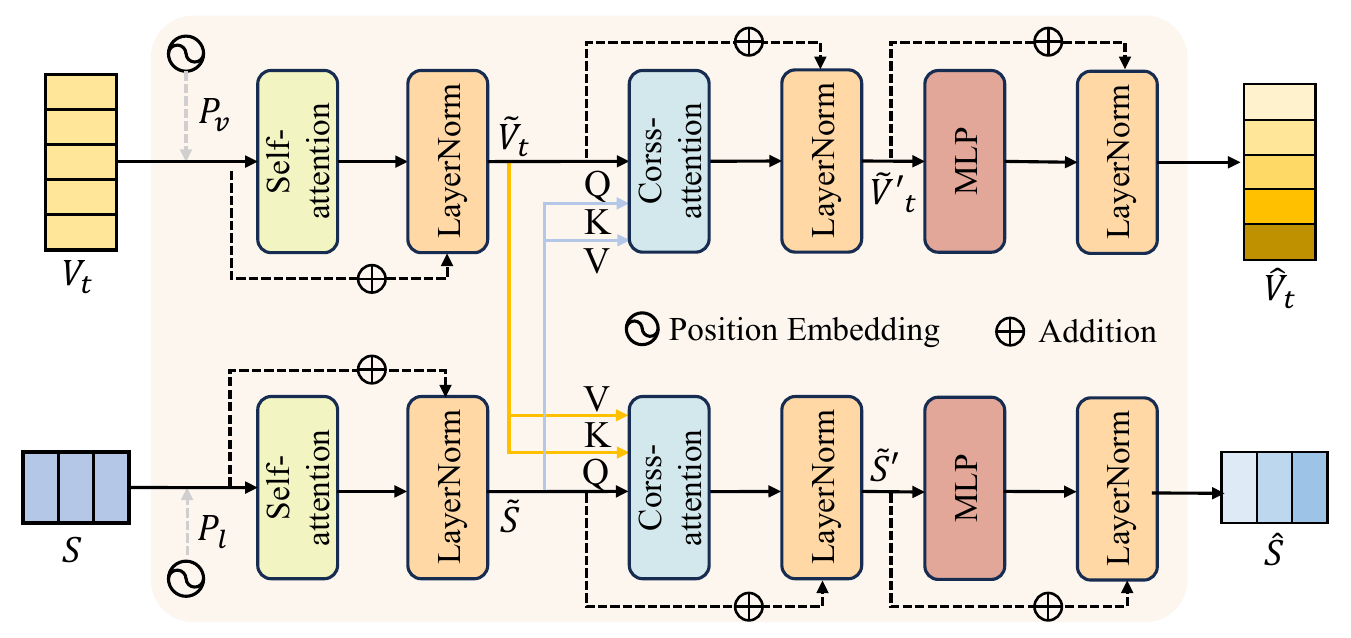}
	\vspace{-0.3cm}
	\caption{
		Overview of the Bidirectional Interactive Fusion module. The"Bidirectional" specifically refers to the cross-modal interaction pattern between visual and language modalities, rather than a forward-backward process.
        }
	\label{fig:BiF}
	\vspace{-0.5cm}
\end{figure*}

\textbf{Bidirectional Interactive Fusion}. A critical challenge in RMOT is effectively aligning language descriptions with visual features while preserving the inherent structural differences between these modalities. Existing methods~\cite{rmot1,rmot2,rmoticassp} typically feed both modality features directly into a multimodal encoder, which often dilutes fine-grained semantic distinctions and spatial relationships. To address this limitation, we propose a Bidirectional Interactive Fusion (BIF) module, as shown in Figure~\ref{fig:BiF}.
By implementing structured interactions before the main encoder, we enable each modality to selectively attend to relevant information from the other while maintaining its inherent characteristics.
Specifically, given visual features $\mathbf{V}_t \in \mathbb{R}^{HW \times C}$ and language embeddings $\mathbf{S} \in \mathbb{R}^{L \times D}$, our encoder first applies modality-specific self-attention with positional encodings:
\begin{equation}
	\begin{aligned}
		\tilde{\mathbf{V}}_t &= \mathbf{V}_t + \text{LN}(\text{SA}(\mathbf{V}_t + \mathbf{P}_v)), \
		\tilde{\mathbf{S}} = \mathbf{S} + \text{LN}(\text{SA}(\mathbf{S} + \mathbf{P}_l)),
	\end{aligned}
\end{equation}
where SA denotes self-attention and LN represents layer normalization. $\mathbf{P}_v$ and $\mathbf{P}_l$ are modality-specific positional encodings. We then model cross-modal interactions through a bidirectional attention mechanism:
\begin{equation}
	\begin{aligned}
		\text{CA}(\mathbf{Q}, \mathbf{K}, \mathbf{V}) &= \text{Softmax}\left(\frac{(\mathbf{Q}W_q)(\mathbf{K}W_k)^T}{\sqrt{d_k}}\right)(\mathbf{V}W_v), \\
		\tilde{\mathbf{V}}'_t &= \tilde{\mathbf{V}}_t + \text{LN}(\text{CA}(\tilde{\mathbf{V}}_t, \tilde{\mathbf{S}}, \tilde{\mathbf{S}})), \\
		\tilde{\mathbf{S}}' &= \tilde{\mathbf{S}} + \text{LN}(\text{CA}(\tilde{\mathbf{S}}, \tilde{\mathbf{V}}_t, \tilde{\mathbf{V}}_t)),
	\end{aligned}
        \label{eq:ca}
\end{equation}
where ${W}_q, {W}_k, {W}_v$ are learnable projection matrices, and CA$(\cdot)$ denotes the cross-attention operation.

The cross-attended features are further refined through feed-forward networks with residual connections:
\begin{equation}
	\begin{aligned}
		\hat{\mathbf{V}}_t &= \tilde{\mathbf{V}}'_t  + \text{LN}(\text{MLP}(\tilde{\mathbf{V}}'_t )) \in \mathbb{R}^{HW\times d}, \
		\hat{\mathbf{S}} = \tilde{\mathbf{S}}' + \text{LN}(\text{MLP}(\tilde{\mathbf{S}}' )) \in \mathbb{R}^{L\times d},
	\end{aligned}
	\label{eq:fusion}
\end{equation}
where $\text{MLP}$ is Multi-Layer Perceptron, and d denotes the model modality-specific dimensions.

This hierarchical structure enables progressive refinement of cross-modal representations while maintaining modality-specific context, facilitating precise alignment between spatial visual features and semantic linguistic elements.
Following the integration of visual and linguistic modalities, we employ a deformable encoder $\text{DeformEnc}(\cdot)$ ~\cite{Detransformer,rmot1,rmot2} to further enhance the language-guided visual features:
\begin{equation} 
\mathbf{F}^v_t = \text{DeformEnc}(\hat{\mathbf{V}}_t) \in \mathbb{R}^{HW \times d}, 
\end{equation} 
the resulting multi-modal representation $\mathbf{F}^v_t $ captures essential correlations between language and visual information, establishing the foundation for object localization and referring expression comprehension in subsequent stages.

\subsection{Progressive Semantic-Decoupled Query Learning}
Most existing RMOT approaches~\cite{rmot1,rmot2,rmoticassp,mls-track,lamot} directly inject language prompts as holistic features into visual features or queries, which overlook the rich semantic information embedded in language descriptions. This plain fusion strategy particularly struggles with complex motion descriptions (\emph{e.g.}, "black cars in the right"), where models often fail to accurately comprehend and localize objects. Inspired by the dual-stream hypothesis in human visual cognition~\cite{brain1,brain3,brain4}, we propose to decouple object perception and context awareness for better language-guided multi-object tracking. Specifically, we employ a linguistic parser~\cite{decouple2,decoupled} that analyzes the syntactic structure of the referring expression to separate static object attributes from spatial-motion information. For instance, noun phrases and adjective phrases that describe object appearance (\emph{e.g.}, "black car," "woman") are extracted as static object information, while prepositional phrases, directional terms, and motion verbs (\emph{e.g.}, "on the left," "moving away,") are identified as spatial-motion information. Our decoupling process implements a linguistically-informed mapping function based on part-of-speech (POS) analysis and dependency parsing. For semantically ambiguous constructions, we apply hierarchical syntactic constraints that preserve phrasal integrity while ensuring mutually exclusive assignments between the two information streams. This decoupling not only aligns with human visual cognitive mechanisms but also enables hierarchical object localization.
We propose a semantic decomposition mechanism to effectively parse the input language expression $\mathcal{T}$ into two complementary streams:
\begin{equation}
		{\mathcal{T}_{so}, \mathcal{T}_{sm}} = \mathbf{Dec}(\mathcal{T}),
\end{equation}
where $\mathcal{T}_{so}$ encodes static object information while $\mathcal{T}_{sm}$ captures spatial motion information, $\mathbf{Dec}(\cdot)$ denotes external linguistic parser~\cite{decoupled} as systematically identify and categorize different parts of speech within the expression.
	
For instance, given the expression ``black cars in the right'', the parser would decompose it into static cues $\mathcal{T}_{so}$: ``black cars'' focusing on static object attributes, and spatial motion information $\mathcal{T}_{sm}$: ``in the right'' describing spatial relationships. Moreover, to maintain the holistic semantic understanding, we augment both feature streams by incorporating the enhanced sentence embedding $\hat{\mathbf{S}}$, which provides essential contextual grounding for the decomposed components. It enables our model to process different semantic aspects more effectively while preserving their inherent relationships within the original expression. Formally:
	
\begin{equation}
		\begin{aligned}
			\mathbf{f}_{so} &= \mathcal{F}_{\text{lan}}(\mathcal{T}_{so}) \in \mathbb{R}^{K_{o} \times D}, \\
			\mathbf{f}_{sm} &= \mathcal{F}_{\text{lan}}(\mathcal{T}_{sm}) \in \mathbb{R}^{K_{m} \times D},
		\end{aligned}
\end{equation}
\begin{equation}
        \begin{aligned}
            \hat{\mathbf{f}}_{so} &= \text{LN}([\mathbf{f}_{so}+ {r}(\hat{\mathbf{S}}); \mathbf{f}_{so}]) \in \mathbb{R}^{K_{o} \times d}, \\
            \hat{\mathbf{f}}_{sm} &= \text{LN}([\mathbf{f}_{sm} + {r}(\hat{\mathbf{S}}); \mathbf{f}_{sm}] \in \mathbb{R}^{K_{m} \times d},
        \end{aligned}
\end{equation}
where $\mathcal{F}_{\text{lan}}(\cdot)$ denotes a pre-trained language model~\cite{roberta}, $K_{o}$ and $K_{m}$ represent the lengths of different attributes in semantics respectively, $D$ and $d$ represent the original text encoder output features dimensions and model modality-specific dimensions, respectively, and $r(\cdot)$ denotes the reshape operation, $[\cdot;\cdot]$ indicates feature concatenation in channel dimension, and $\text{LN}(\cdot)$ applies layer normalization to stabilize feature distributions. 
	
\begin{figure*}[!t]
		\centering
		\includegraphics[width=0.8\linewidth]{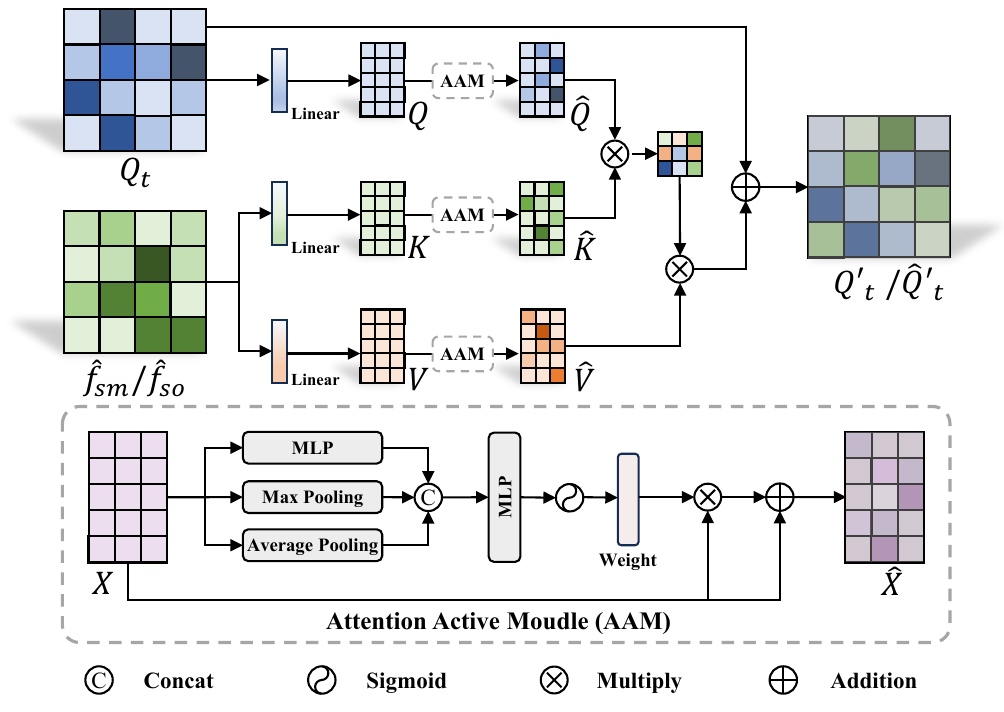}
		\vspace{-0.3cm}
		\caption{
			Overview of the Query Semantic Injection (QSI) module architecture. It efficiently incorporates disentangled semantic features ($\hat{\mathbf{f}}_{so}$ for static object attributes or $\hat{\mathbf{f}}_{sm}$ for spatial motion information) into query representations $Q_t$. The Attention Active Module (AAM) dynamically enhances feature saliency through multi-perspective feature aggregation, generating adaptive attention weights that modulate the cross-attention process. This enables queries to selectively assimilate significant information, facilitating fine-grained semantic-aware object localization.
		}
		\label{fig:QSI}
		\vspace{-0.5cm}
\end{figure*}
	
Current RMOT methods primarily focus on fusing visual and language features at the encoder stage ~\cite{rmot1,rmot2,rmoticassp}, where textual information is directly integrated into visual features through cross-modal attention. While this early fusion strategy achieves basic semantic alignment, it faces two key limitations. First, the rich hierarchical nature of language semantics is not fully leveraged. Second, queries, serving as learnable object prototypes for trajectory modeling, lack explicit semantic guidance in their initialization phase. We argue that comprehensive semantic understanding and queries play a pivotal role in RMOT as they directly influence object localization and temporal association. Unlike conventional object queries that only encode appearance patterns, queries in RMOT need to be semantically aware to distinguish referred objects from other visible objects. Therefore, we propose a progressive query learning mechanism that injects semantic prior information directly into queries before the decoder stage, enabling them to carry explicit language guidance throughout the detection process.
	
Specifically, our approach integrated semantic information into detection and tracking query sets $\mathbf{Q}_t = \{{\mathbf{Q}^{Det}_t, \mathbf{Q}^{Tra}_t}\}$ through progressive learning. It creates a synergistic relationship between linguistic cues and visual representations, circumventing potential semantic dilution during feature propagation and enabling fine-grained object localization through semantic pattern matching. 
Our decoder architecture comprises $L$ transformer layers structured hierarchically. The initial $L/2$ layers establish foundational object perception capabilities, while the subsequent layers develop sophisticated contextual understanding. To formalize this process, we introduce the Query Semantic Injection (QSI) Module, which incorporates semantic features into query representations during the transformer decoding phase, as shown in Figure~\ref{fig:QSI}. This enables direct semantic conditioning of object queries, which serve as learnable object prototypes crucial for trajectory modeling in tracking tasks. To maximize the utilization of important information in each vector, we employ the Attention Active Module (AAM), which implements multi-perspective feature aggregation by simultaneously utilizing direct feature transformation alongside complementary statistical pooling operations.  It allows the model to dynamically modulate attention weights between query features and semantic embeddings, effectively enhancing feature saliency in a context-aware manner.
QSI first enhances queries $\mathbf{Q}_t$ by integrating static semantic features $\hat{\mathbf{f}}_{so}$ through the following formulation:
\begin{equation}
		\begin{aligned}
			AAM(\textbf{X}) &= \textbf{X} + \textbf{X}\mathcal{W}(\textbf{X}),\\
			\mathcal{W}(\textbf{X})
			&= \sigma\Big(\Psi_1(\text{Concat}[\Psi_2(\textbf{X}), \text{Max}(\textbf{X}), \text{Avg}(\textbf{X})]\big)\Big),
		\end{aligned}
\end{equation}
\begin{equation}
		\begin{aligned}
			\mathbf{Q}^{so}_{t} &= \text{QSI}(\mathbf{Q}_t, \hat{\mathbf{f}}_{so}), \\
			&= \mathbf{Q}_t + \frac{AAM{(\mathbf{Q}_t)}AAM{(\hat{\mathbf{f}}_{so}})^T}{\sqrt{d}}AAM{(\hat{\mathbf{f}}_{so})},\\
		\end{aligned}
\end{equation}
where $\hat{\mathbf{f}}_{so}$ denotes fundamental object characteristics, including appearance, category, and chromatic properties. $\Psi_1$ and $\Psi_2$ represent feature transformation and fusion MLPs respectively, $\sigma$ denotes the sigmoid function and $\text{LN}(\cdot)$ is the layer normalization function. 
The resulting queries $\hat{\mathbf{Q}}^{so}_{t}$ are then fed into decoder layers ($\text{DecLayers}(\cdot)$) for interaction with visual features:
\begin{equation}
		\begin{aligned}
			\mathbf{Q}'_t &= \text{DecLayers}_{0{\rightarrow}L/2}(\mathbf{Q}^{so}_{t},\mathbf{F}^v_t ),
		\end{aligned}
\end{equation}
where $L$ represents the total number of decoder layers in our architecture.

This initial injection helps queries establish basic object recognition capabilities aligned with the referring expression.
Building upon the enhanced queries $\mathbf{Q}'_t$, we incorporate spatial motion semantic features $\hat{\mathbf{f}}_{sm}$ for capturing motion patterns and spatial relationships. Finally, we regress discriminative instance queries $\hat{\mathbf{Q}}'$ for the referred tracks in current frame  $\mathbf{F}_t$:
\begin{equation}
		\begin{aligned}
			\hat{\mathbf{Q}}'_t &= \text{DecoderLayers}_{L/2{\rightarrow}L}(\text{QSI}(Q', \hat{\mathbf{f}}_{sm}),\mathbf{F}^v_t ).
		\end{aligned}
\end{equation}
	
Progressive Semantic-Decoupled Query Learning (PSDQL) employs a progressive semantic transition from basic to complex representation, allowing queries to develop a comprehensive understanding of referred objects.  
\subsection{Structural Consistency Constraint}
Referring Multi-Object Tracking (RMOT) aims to localize and track specific objects in video sequences based on language references. While existing RMOT methods~\cite{rmot1,rmot2,rmot-tim} have shown promising results by leveraging cross-modal features, they often struggle in challenging scenarios with multiple similar objects or complex linguistic descriptions. This misalignment stems from a fundamental challenge: the mapping between visual objects and language descriptions is inherently many-to-many—the same object can be described differently from various perspectives (\emph{e.g.}, "left cars in light color" or "left cars in silver"), while similar descriptions might refer to different objects. Previous works typically do not fully capture these complex semantic relationships and lack effective mechanisms to verify whether tracked objects truly correspond semantically to the referring expressions.
\begin{figure*}[!t]
	\centering
	\includegraphics[width=\linewidth]{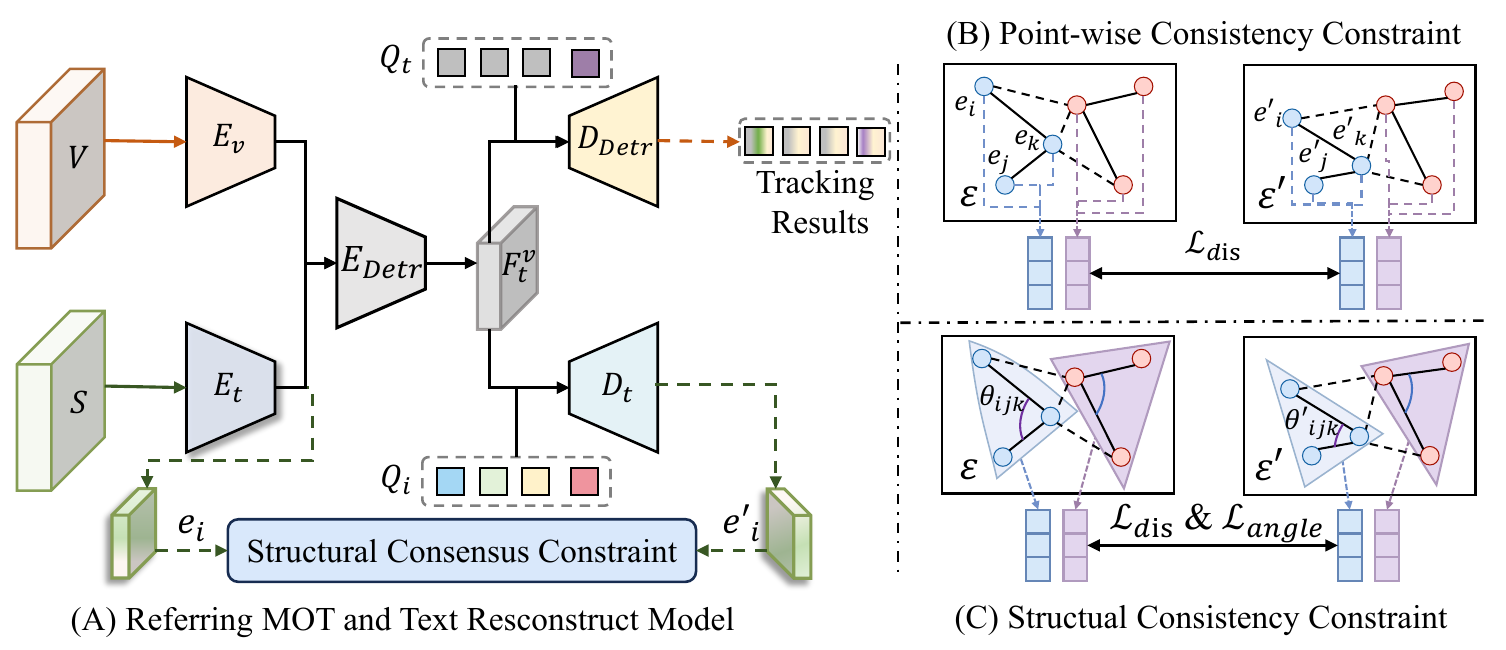}
	\vspace{-0.1cm}
	\caption{
		(A) Overview of our Referring Multi-Object Tracking model with structural consistency constraints. (B) Point-wise consistency constraint enforces direct correspondence between entity embeddings ($e_i$ and $e'_i$) through distance-based matching ($\mathcal{L}_{dis}$). (C) Our structural consistency constraint preserves geometric relationships between entities in both embedding spaces ($\mathcal{E}$ and $\mathcal{E}'$) using both distance and angle consistency losses ($\mathcal{L}_{dis}$ \& $\mathcal{L}_{angle}$), ensuring that semantic relationships between objects remain consistent even when specific descriptions vary.
        }
	\label{fig:SCC}
	\vspace{-0.5cm}
\end{figure*}

In this work, we address these challenges by introducing a structural consistency constraint that bridges visual-language understanding in both forward and backward directions, as shown in Figure~\ref{fig:SCC}. Applied exclusively during the training stage, it serves as a powerful regularization mechanism without adding inference overhead. Specifically, we leverage our previously disentangled textual information to obtain representations for different semantic components through Equation.~\ref{eq:text}. This process yields three corresponding embeddings: static object embedding $\mathbf{e}_{so} \in \mathbb{R}^{1\times C}$, spatial motion embedding $\mathbf{e}_{sm} \in \mathbb{R}^{1\times C}$, and the global sentence representation $\mathbf{e}_g \in \mathbb{R}^{1\times C}$.
Through cross-modal interaction, we derive the corresponding visual proxy features. The cross-attention mechanism facilitating this interaction is formulated as:
	
\begin{equation}
		\begin{aligned}
			\mathbf{F}^{so}_t &= \tilde{\mathbf{V}}_t + \text{LN}(\text{CA}(\tilde{\mathbf{V}}_t, \mathbf{S}_{so},\mathbf{S}_{so})), \\
			\mathbf{F}^{sm}_t &= \tilde{\mathbf{V}}_t + \text{LN}(\text{CA}(\tilde{\mathbf{V}}_t, \mathbf{S}_{sm}, \mathbf{S}_{sm})), \\
		\end{aligned}
\end{equation}
where CA$(\cdot)$ denotes the cross attention mechanisms in Equation.~\ref{eq:ca}.

After obtaining these visual proxy features $\mathbf{F}^v_{so}$, $\mathbf{F}^v_{sm}$, and $\mathbf{F}^v_t$, we employ a text decoder $\mathcal{D}_t$ to project these representations back to the textual embedding space. This visual-to-textual (V-T) projection mechanism~\cite{textreconstruction} utilizes  transformer decoder $\mathcal{D}_t$ that transforms the visual representation of the referred object into the textual space. As illustrated in Figure~\ref{fig:SCC}-A, the learnable text query $\mathbf{Q}_i \in \mathbb{R}^{d \times 1}$ is employed to query the proxy features $\mathbf{F}^{v}_i$. The transformer decoder outputs the reconstructed text embedding $\mathbf{e}'_i$ according to:
\begin{equation}
		\begin{aligned}
			\mathbf{e}'_i &= \mathcal{D}_t(\mathbf{F}^{v}_i, \mathbf{Q}_i) \in \mathbb{R}^{d \times 1}, \\
		\end{aligned}
\end{equation}
where $\mathcal{D}_t(\cdot)$ denotes the transformer text reconstruction decoder.

This reconstruction process enables the model to verify whether the tracked object sequence truly captures the semantic information specified in the original referring expression.
To effectively validate the semantic alignment between tracked objects and their corresponding referring expressions, we need to quantify the consistency between original text embeddings and their reconstructed counterparts. However, establishing an appropriate consistency measure presents significant challenges. The point-wise consistency (Figure~\ref{fig:SCC}-B) constraints widely adopted in previous cross-modal tasks fall short in RMOT scenarios due to a fundamental limitation: they assume a bijective mapping between visual objects and textual descriptions. From an information-theoretic standpoint, a point-wise matching constraint restricts the mutual information between visual and textual modalities. When considering the visual features $\mathbf{F}^v$ and text embeddings $\mathbf{e}$ as random variables, the point-wise matching effectively constrains their joint distribution $p(\mathbf{F}^v, \mathbf{e})$ to be deterministic. However, in real-world RMOT scenarios, the optimal representation should maximize the mutual information $I(\mathbf{F}^v, \mathbf{e})$ while accommodating the inherent many-to-many relationships between modalities:
    \begin{equation}
    I(\mathbf{F}^v, \mathbf{e}) = \sum_{\mathbf{F}^v, \mathbf{e}} p(\mathbf{F}^v, \mathbf{e}) \log \frac{p(\mathbf{F}^v, \mathbf{e})}{p(\mathbf{F}^v)p(\mathbf{e})},
    \end{equation}
point-wise matching between the original expression $\mathcal{E}$ and the reconstructed embedding $\mathcal{E}'$ would artificially constrain this information flow, leading to sub-optimal representations.

To address this limitation, we introduce a structural consistency constraint (Figure~\ref{fig:SCC}-C) that preserves semantic relationships while accommodating language variations. Instead of enforcing strict one-to-one correspondence between expressions, we maintain the relative structure of embedding spaces $\mathcal{E}$ and $\mathcal{E}'$. The key insight is that while the same tracked object may have different valid descriptions, the semantic relationships between different objects should remain consistent. 
Our structural consistency constraint focuses on preserving two essential geometric properties across embedding spaces. Distance preservation ensures that relative distances between different referring expressions remain consistent in both original and reconstructed spaces, maintaining the integrity of semantic similarities. Simultaneously, angle preservation captures higher-order semantic structure by maintaining geometric relationships between expression triplets. When tracked objects correctly match their referring expressions, these structural properties remain well-preserved between $\mathcal{E}$ and $\mathcal{E}'$, while mismatched trajectories manifest as structural inconsistencies. This approach allows our model to accommodate linguistic diversity while maintaining semantic consistency, significantly enhancing the robustness of multi-object tracking in language-guided scenarios.
Formally, we define the structural consistency loss as:
	\begin{equation}
		\mathcal{L}_{\text{struct}} = \mathcal{L}_{\text{dist}} + \lambda_{angle}\mathcal{L}_{\text{angle}},
	\end{equation}
where $\lambda_{angle}$ = 0.4 is the balance hyperparameter, $\mathcal{L}_{\text{dist}}$ measures the preservation of pairwise distances and $\mathcal{L}_{\text{angle}}$ evaluates the consistency of geometric relationships between expression triplets. 
	
For a set of expression pairs $(e_i, e_j)$ in the original space and their reconstructed counterparts $(e'_i, e'_j)$, the distance preservation loss is computed as:
\begin{equation}
		\mathcal{L}_{\text{dist}} = \sum_{(e_i, e_j) \in \mathcal{P}_2} l_{h}\left(\frac{||e_i - e_j||_2}{\bar{d}} - \frac{||e'_i - e'_j||_2}{\bar{d}'}\right),
\end{equation}
where $\mathcal{P}_2$ represents the set of all possible embedding pairs from the original and reconstructed spaces such that $i \neq j$, $||e_i - e_j||_2$ denotes the Euclidean distance between embeddings $e_i$ and $e_j$, $\bar{d}$ is the average distance across all embedding pairs in the original space defined as $\bar{d} = \frac{1}{|\mathcal{P}_2|}\sum_{(e_i, e_j) \in \mathcal{P}_2} ||e_i - e_j||_2$, and $\bar{d}'$ is the corresponding average in the reconstructed space.
	
Similarly, the angle preservation loss is defined as:
\begin{equation}
		\mathcal{L}_{\text{angle}} = \sum_{(e_i, e_j, e_k) \in \mathcal{P}_3} l_{h}\left(\cos\angle(e_i, e_j, e_k) - \cos\angle(e'_i, e'_j, e'_k)\right),
\end{equation}
where $\mathcal{P}_3$ denotes the set of all possible embedding triplets from the original and reconstructed spaces such that $i \neq j \neq k$, $\cos\angle(e_i, e_j, e_k)$ represents the cosine of the angle formed by the vectors from $e_i$ to $e_j$ and from $e_i$ to $e_k$, calculated as:
\begin{equation}
		\cos\angle(e_i, e_j, e_k) = \frac{(e_j-e_i) \cdot (e_k-e_i)}{||e_j-e_i||_2 \cdot ||e_k-e_i||_2}.
\end{equation}

Both components utilize the Huber loss function $l_{h}$, which provides robustness against outliers while maintaining sensitivity to small deviations:
\begin{equation}
l_{h}(x) = \begin{cases} 
\frac{1}{2}x^2, & \text{if}\ |x| \leq 1 \\
|x| - \frac{1}{2}, & \text{otherwise}
\end{cases}
\end{equation}

\subsection{Loss Functions}
Our framework is jointly optimized with a combination of detection objectives and cross-modal structural learning. For object classification, we adopt focal loss $\mathcal{L}_{\text{cls}}$ to address class imbalance. Accurate bounding box localization is achieved through complementary regression terms: $\mathcal{L}_{\text{L1}}$ for coordinate-wise accuracy and $\mathcal{L}_{\text{giou}}$ for overall box IoU optimization. The referring correspondence between visual objects and linguistic descriptions is supervised by $\mathcal{L}_{\text{ref}}$, while bidirectional semantic consistency between visual and language embeddings is enforced through $\mathcal{L}_{\text{struct}}$. The overall loss function is formulated as:
	\begin{equation}
		\mathcal{L} = \lambda_{\text{cls}}\mathcal{L}_{\text{cls}} + \lambda_{\text{L1}}\mathcal{L}_{\text{L1}} + \lambda_{\text{giou}}\mathcal{L}_{\text{giou}} + \lambda_{\text{ref}}\mathcal{L}_{\text{ref}} + \lambda_{\text{struct}}\mathcal{L}_{\text{struct}},
	\end{equation}
where $\lambda{(\cdot)}$ are the balancing weights for training and following the same settings as in~\cite{rmot1,rmot2}.
\begin{algorithm}
        \caption{Cognitive Disentanglement for Referring Multi-Object Tracking}
        \begin{algorithmic}[1]
        \REQUIRE Video frames $\{\mathbf{F}_t\}_{t=1}^T$, language description $\mathcal{T}$
        \ENSURE Tracking results for referred objects
        
        \STATE $\mathbf{S} \leftarrow E_l(\mathcal{T})$,  \COMMENT{Language encoding}
        \STATE $\{\mathcal{T}_{so}, \mathcal{T}_{sm}\} \leftarrow \text{Dec}(\mathcal{T})$ \COMMENT{Semantic decoupling}
        \FOR{$t = 1$ to $T$}
            \STATE $\mathbf{V}_t \leftarrow E_v(\mathbf{F}_t)$ \COMMENT{Visual encoding}
            
            \STATE \textbf{// Bidirectional Interactive Fusion}
            \STATE $\hat{\mathbf{V}}_t, \hat{\mathbf{S}} \leftarrow \text{BIF}(\mathbf{V}_t, \mathbf{S})$ \COMMENT{Cross-modal fusion}
            \STATE $\mathbf{F}^v_t \leftarrow \text{DeformEnc}(\hat{\mathbf{V}}_t) \in \mathbb{R}^{HW \times d}$
            \STATE $\hat{\mathbf{f}}_{so}, \hat{\mathbf{f}}_{sm} \leftarrow \text{ProcessTextFeatures}(\mathcal{T}_{so}, \mathcal{T}_{sm}, \hat{\mathbf{S}})$
            \STATE \textbf{// Progressive Semantic-Decoupled Query Learning}
            \STATE $\mathbf{Q}^{so}_t \leftarrow \text{QSI}(\mathbf{Q}_t, \hat{\mathbf{f}}_{so})$ \COMMENT{What pathway}
            \STATE $\mathbf{Q}'_t \leftarrow \text{DecLayers}_{0 \rightarrow L/2}(\mathbf{Q}^{so}_t, \mathbf{F}^v_t)$
            \STATE $\hat{\mathbf{Q}}'_t \leftarrow \text{DecLayers}_{L/2 \rightarrow L}(\text{QSI}(\mathbf{Q}'_t, \hat{\mathbf{f}}_{sm}), \mathbf{F}^v_t)$ \COMMENT{Where pathway}
            \STATE $\mathbf{b}_t, \mathbf{c}_t, \mathbf{r}_t \leftarrow \text{PredictionHeads}(\hat{\mathbf{Q}}'_t)$
            \STATE Association and tracking management if $t > 1$
            \STATE Update queries for next frame
            \IF{Training}
                \STATE \textbf{// Structural Consistency Constraint}
                \STATE $\mathbf{e}'_{so}, \mathbf{e}'_{sm}, \mathbf{e}'_g \leftarrow \text{TextReconstruction}(\mathbf{F}^v_t, \mathbf{F}^{so}_t, \mathbf{F}^{sm}_t)$
                \STATE $\mathcal{L}_{struct} \leftarrow \text{StructuralLoss}(\mathbf{e}_{so}, \mathbf{e}_{sm}, \mathbf{e}_g, \mathbf{e}'_{so}, \mathbf{e}'_{sm}, \mathbf{e}'_g)$
                \STATE \textbf{// RMOT losses}
                \STATE $\mathcal{L}^{det}_t \leftarrow \sum^{N}_{i=1}[\lambda_{cls}\mathcal{L}_{cls}(\mathbf{c}^{det}_{t,i}, \hat{\mathbf{c}}^{det}_{t,i}) + \mathbb{1}_{\{\hat{\mathbf{c}}^{det}_{t,i} \neq \emptyset\}}(\mathcal{L}_{box}(\mathbf{b}^{det}_{t,i}, \hat{\mathbf{b}}^{det}_{t,i}) + \lambda_{ref}\mathcal{L}_{ref}(\mathbf{r}^{det}_{t,i}, \hat{\mathbf{r}}^{det}_{t,i}))]$
                \STATE $\mathcal{L}^{tra}_t \leftarrow \sum^{N'_{t-1}}_{i=1}[\lambda_{cls}\mathcal{L}_{cls}+ \mathcal{L}_{box} + \lambda_{ref}\mathcal{L}_{ref}]$
                \IF{$t = 1$}
                     \STATE $\mathcal{L}_t \leftarrow  \mathcal{L}^{det}_t + \lambda_{struct}\mathcal{L}_{struct}$
                \ELSE
                    \STATE $\mathcal{L}_t \leftarrow \mathcal{L}^{tra}_t + \mathcal{L}^{det}_t + \lambda_{struct}\mathcal{L}_{struct}$
                \ENDIF    
            \ENDIF
        \ENDFOR
        \end{algorithmic}
     \end{algorithm}
\section{Experiments}
\label{sec:exp}
\subsection{Implementation Details}
We adopt TransRMOT as our baseline architecture, which employs ResNet-50~\cite{resnet50} as the visual backbone and RoBERTa~\cite{roberta} as the text encoder. The text encoder's parameters are kept frozen throughout the training process to maintain stable linguistic representations.
The model is initialized with Deformable DETR~\cite{Detransformer} weights pre-trained on the COCO dataset~\cite{coco}, except for the text encoder. We set the number of detection queries to $N = 300$ to ensure sufficient coverage of potential objects. 
The network is trained for 100 epochs on 2 NVIDIA RTX A5000 24GB GPUs with a batch size of 1 per GPU. We utilize AdamW~\cite{adamw} optimizer with a base learning rate of $1e^{-4}$, while the backbone's learning rate is set to $1e^{-5}$. A learning rate decay of 0.1 is applied at the 50th epoch. For data augmentation, we employ random cropping and incorporate object erasing and inserting operations to enhance robustness to object entrances and exits, following~\cite{motr}. 
The loss function is weighted with coefficients $\lambda_{cls} = 2$, $\lambda_{L1} = 5$, $\lambda_{giou} = 2$, $\lambda_{ref} = 2$, and $\lambda_{struct} = 2$  to balance different training objectives.

During inference, CDRMT processes video sequences of arbitrary length in an online manner without requiring post-processing. At frame $t$, the model generates $N_t$ instance embeddings, each corresponding to either a true object or an empty prediction. We employ a two-stage filtering strategy: first, objects with confidence scores exceeding 0.7 are considered as valid detections; subsequently, the referring threshold $\beta_{ref} = 0.5$ is applied to identify the final referred objects from these valid detections.
The entire pipeline of the method (CDRMT) is detailed in Algorithm 1.
\subsection{Datasets and Evaluation Metrics}
\subsubsection{Datasets}We conduct extensive experiments on two challenging RMOT benchmarks: Refer-KITTI~\cite{rmot1}, and its extended version Refer-KITTI-V2~\cite{rmot2}. 

Refer-KITTI~\cite{rmot1} contains 18 high-resolution videos with 818 natural language expressions. Each expression corresponds to an average of 10.7 objects, with temporal spans ranging from 0 to 400 frames. For evaluation on this dataset, we address a critical timing synchronization issue in the original benchmark where frame predictions were misaligned with ground truth annotations. 

Refer-KITTI-V2~\cite{rmot2} extends the original benchmark by utilizing all 21 videos from KITTI, introducing more challenging scenarios with sequences up to 1,059 frames. This version features more diverse expressions describing both motion patterns and visual attributes. To facilitate comprehensive evaluation, we develop enhanced baselines by integrating cross-modal fusion capabilities into established tracking frameworks like FairMOT~\cite{fairmot} and ByteTrack~\cite{bytetrack}.
\begin{table}[t]
	\centering
	\setlength{\tabcolsep}{2.5pt} 
	\renewcommand{\arraystretch}{1.4} 
	\caption{Comparison with state-of-the-art methods on Refer-KITTI dataset. $\uparrow$/ indicates higher scores are better. The best results are highlighted in bold. * means the result after frame correction strategy.}
	\label{tab:comparison-kittiv1}
	\resizebox{\textwidth}{!}{
		\begin{tabular}{l|c|c|ccccccccc}
			\toprule[1.5pt]
			Method & Backbone & Detector & HOTA$\uparrow$ & DetA$\uparrow$ & AssA$\uparrow$ & DetRe$\uparrow$ & DetPr$\uparrow$ & AssRe$\uparrow$ & AssPr$\uparrow$ & LocA$\uparrow$ \\
			\midrule[1pt]
			DeepSORT\cite{deepsort}$_{\text{ICIP17}}$  & - & FairMOT & 25.59 & 19.76 & 34.31 & 26.38 & 36.93 & 39.55 & 61.05 & 71.34 \\
			FairMOT\cite{fairmot}$_{\text{IJCV21}}$  & DLA-34 & CenterNet & 23.46 & 14.84 & 40.15 & 16.44 & 45.84 & 43.05 & 71.65 & 74.77 \\
			ByteTrack\cite{bytetrack}$_{\text{ECCV22}}$  & - & FairMOT & 24.95 & 15.50 & 43.11 & 18.25 & 43.48 & 48.64 & 70.72 & 73.90 \\
			CSTrack\cite{rethinking}$_{\text{TIP22}}$  & DarkNet-53 & YOLOv5 & 27.91 & 20.65 & 39.00 & 33.76 & 32.61 & 43.12 & 71.82 & 79.51 \\
			MOTRv2\cite{motrv2}$_{\text{CVPR23}}$  & ResNet-50 & YOLOX+DAB-D-DETR & 37.56 & 25.49 & 56.90 & - & - & - & - & - \\
			MGLT-MOTRv2\cite{rmot-tim}$_{\text{TIM25}}$  & ResNet-50 & YOLOX+DAB-D-DETR & 39.07 & 26.74 & 58.37 & - & - & - & - & - \\
			\midrule
			TransTrack\cite{TransT}$_{\text{arXiv20}}$  & ResNet-50 & Deformable-DETR & 32.77 & 23.31 & 45.71 & 32.33 & 42.23 & 49.99 & 78.74 & 79.48 \\
			TrackFormer\cite{MOT1}$_{\text{CVPR22}}$  & ResNet-50 & Deformable-DETR & 33.26 & 25.44 & 45.87 & 35.21 & 42.10 & 50.26 & 78.92 & 79.63 \\
			CO-MOT\cite{bridging}$_{\text{arXiv23}}$  & ResNet-50 & Deformable-DETR & 36.95 & 25.80 & 54.52 & - & - & - & - & - \\
			TransRMOT\cite{rmot1}$_{\text{CVPR23}}$  & ResNet-50 & Deformable-DETR & 38.06 & 29.28 & 50.83 & - & - & - & - & - \\
			EchoTrack\cite{echotrack}$_{\text{TITS24}}$ & ResNet-50 & Deformable-DETR & 39.47 & $31.19$ & 51.56 & 42.65 & 48.86 & 56.68 & 81.21 & 79.93 \\
			DeepRMOT\cite{rmoticassp}$_{\text{ICASSP24}}$  & ResNet-50 & Deformable-DETR & 39.55 & 30.12 & 53.23 & 41.91 & 47.47 & 58.47 & 82.16 & 80.49 \\
			MGLT-MOTR\cite{rmot-tim}$_{\text{TIM25}}$  & ResNet-50 & Deformable-DETR & 39.91 & 31.56 & 51.79 & - & - & - & - & - \\
			MGLT-CO-MOT\cite{rmot-tim}$_{\text{TIM25}}$  & ResNet-50 & Deformable-DETR & 40.79 & 28.91 & 58.78 & - & - & - & - & - \\
			CDRMT (ours) & ResNet-50 & Deformable-DETR & 40.91 & 32.04 & 53.61 & 40.94 & 49.30 & 46.97 & 82.43 & 80.64 \\
			\midrule
			TransRMOT\cite{rmot1}*$_{\text{CVPR23}}$  & ResNet-50 & Deformable-DETR & 46.56 & 37.97 & 57.33 & 49.69 & \textbf{60.10} & 60.02 & 89.67 & 90.33 \\
			CO-MOT\cite{bridging}*$_{\text{arXiv23}}$  & ResNet-50 & Deformable-DETR & 45.22 & 33.90 & 60.45 & - & - & - & - & - \\
			iKUN\cite{ikun}*$_{\text{CVPR24}}$  & ResNet-50 & Deformable-DETR & 48.84 & 35.74 & \textbf{66.80} & 51.97 & 52.25 & \textbf{72.95} & 87.09 & - \\
			MLS-Track\cite{mls-track}*$_{\text{arXiv24}}$  & ResNet-50 & Deformable-DETR & $49.05$ & $40.03$ & 60.25 & \textbf{59.07} & 54.18 & 65.12 & 88.12 & - \\
			MGLT-MOTR\cite{rmot-tim}$_{\text{TIM25}}$* & ResNet-50 & Deformable-DETR & 47.95 & 40.04 & 57.57 & - & - & - & - & - \\
			MGLT-CO-MOT\cite{rmot-tim}*$_{\text{TIM25}}$ & ResNet-50 & Deformable-DETR & 49.25 & 37.09 & 65.50 & - & - & -&-&- \\
			CDRMT* (ours) & ResNet-50 & Deformable-DETR & \textbf{49.35} & \textbf{40.34} & 60.56 & 54.54 & 59.30 & 64.70 & \textbf{89.80} & \textbf{90.61} \\
			\bottomrule[1.5pt]
		\end{tabular}
	}
\end{table}

\subsubsection{Evaluation Metrics}We adopt the Higher Order Tracking Accuracy (HOTA) as our primary evaluation metric, which provides a balanced assessment of both detection and association performance through their geometric mean:
\begin{equation}
	\mathrm{HOTA} = \sqrt{\mathrm{DetA}\cdot \mathrm{AssA}}
\end{equation}
where $\mathrm{DetA}$ quantifies detection accuracy through IoU scores, and $\mathrm{AssA}$ measures the quality of temporal associations. 

\subsection{Comparison With State-of-the-Art Methods}
\subsubsection{Evaluation on Refer-KITTI}
The experimental results in Table~\ref{tab:comparison-kittiv1} demonstrate that our proposed CDRMT achieves new state-of-the-art performance on the challenging Refer-KITTI dataset. Through comprehensive evaluation across multiple metrics, CDRMT exhibits superior performance in both detection precision and tracking robustness.

In terms of overall tracking performance, CDRMT achieves the highest HOTA score of 49.35, surpassing our baseline TransRMOT* (46.56) by 6.0\% and outperforming other competitive methods like iKUN* (48.84) and MLS-Track (49.05). This improvement demonstrates the effectiveness of our method in enhancing both detection and association capabilities.
For detection-specific metrics, CDRMT demonstrates remarkable performance improvements over the TransRMOT* baseline. Specifically, our method achieves a DetA score of 40.34 (improved by 6.2\% from 37.97) and a DetRe score of 54.54 (improved by 9.8\% from 49.69). The DetPr score shows minimal difference (59.30 versus 60.10).
CDRMT shows competitive performance with an AssRe score of 64.70 (improved by 7.8\% from TransRMOT*'s 60.02) and a slight AssPr score of 89.80 (improved by 0.1\% from 89.67). These improvements validate our method's enhanced capability in maintaining consistent object identities through language-guided association.
Furthermore, CDRMT achieves state-of-the-art performance in spatial accuracy with a LocA score of 90.61, showing a slight but consistent improvement (0.3\%) over TransRMOT*'s 90.33. This demonstrates our method's ability to maintain precise object localization while enhancing overall tracking performance.
When compared with other recent state-of-the-art methods, CDRMT shows comprehensive advantages. Notably, it outperforms the strong competitor iKUN* in multiple aspects, with improvements in HOTA (1.0\%), DetA (12.9\%), and association metrics. These consistent improvements across various metrics validate the effectiveness and robustness of our proposed approach in handling the challenging RMOT task.
\subsubsection{Evaluation on Refer-KITTI-V2}
As shown in Table~\ref{tab:comparison-kittiv2}, we conduct comprehensive experiments to evaluate CDRMT against state-of-the-art methods on the Refer-KITTI-V2 dataset. Our comparisons include both conventional MOT methods adapted for referring tracking (FairMOT~\cite{fairmot}, ByteTrack~\cite{bytetrack}) and specialized RMOT approaches (iKUN~\cite{ikun}, TransRMOT~\cite{rmot1}). CDRMT demonstrates consistent superiority across all evaluation metrics, establishing new state-of-the-art performance on this benchmark.
\begin{table}
	\renewcommand{\arraystretch}{1.4} 
	\caption{Comparison with state-of-the-art methods on Refer-KITTI-V2 dataset. The best results are in bold.}
	\label{tab:comparison-kittiv2}
	\begin{tabular*}{\textwidth}{@{\extracolsep{\fill}}l|cccccccc@{}}
		\toprule[1.5pt]
		Method   & HOTA$\uparrow$ & DetA$\uparrow$ & AssA$\uparrow$ & DetRe$\uparrow$ & DetPr$\uparrow$ & AssRe$\uparrow$ & AssPr$\uparrow$ & LocA$\uparrow$ \\
		\midrule[1pt]
		FairMOT~\cite{fairmot} &  22.53 & 15.80 & 32.82 & 20.60 & 37.03 & 36.21 & 71.94 & 78.28 \\
		ByteTrack~\cite{bytetrack} & 24.59 & 16.78 & 36.63 & 22.60 & 36.18 & 41.00 & 69.63 & 78.00 \\
		iKUN~\cite{ikun} &10.32 & 2.17 & 49.77 & 2.36 & 19.75 & \textbf{58.48} & 68.64 & 74.56 \\
		\midrule
		TransRMOT~\cite{rmot1} &  31.00 & 19.40 & 49.68 & \textbf{36.41} & 28.97 & 54.59 & 82.29 & 89.82 \\
		CDRMT(Ours) &\textbf{31.99} & \textbf{20.37} & \textbf{50.35} & 26.40 & \textbf{46.26} & 53.40 &\textbf{85.90} & \textbf{90.36} \\
		\bottomrule[1.5pt]
	\end{tabular*}
\end{table}
Particularly noteworthy is our method's substantial improvement in precision-related metrics. CDRMT achieves the DetPr of 46.26, significantly outperforming TransRMOT by 28.97 and more than doubling the performance of iKUN (19.75). This remarkable enhancement in detection precision highlights the effectiveness of our progressive query injection mechanism in accurately localizing object objects. Similarly, our approach shows strong performance in association-related metrics, with an AssA of 50.35 and an AssPr of 85.90, surpassing TransRMOT by 1.3\% and 4.4\% respectively. The holistic tracking performance, measured by HOTA (31.99) and DetA (20.37), also demonstrates marginal but consistent improvements over the previous best method TransRMOT. Furthermore, our method achieves the highest LocA score of 90.36, indicating superior localization accuracy. These comprehensive improvements across various metrics validate the effectiveness of our cognitive-inspired dual-stream architecture in handling complex referring expressions and diverse tracking scenarios.

\subsubsection{Evaluation on Model Efficiency }
Table~\ref{tab:efficiency} presents a comprehensive efficiency comparison of our CDRMT with state-of-the-art RMOT methods. For fair comparison, all methods adopt Deformable-DETR as the base detector with identical backbone architecture, ensuring that performance differences stem from the referring tracking modules rather than detection capabilities. Despite achieving superior tracking performance, CDRMT maintains competitive computational efficiency with an inference speed of 10.11 FPS.

Specifically, compared to TransRMOT, our method requires comparable parameters (126.0M vs 124.16M) while maintaining competitive inference speed (10.11 FPS vs 10.38 FPS). In terms of computational complexity, CDRMT demonstrates significant advantages over recent methods like iKUN and TempRMOT, reducing FLOPs by 42.2\% (287.03G vs 496.98G) and 51.2\% (287.03G vs 588.70G) respectively. This efficiency gain can be attributed to our cognitive-inspired architecture that eliminates redundant feature processing while maintaining effective visual-linguistic reasoning.
While MGLT-CO-MOT achieves lightweight design with only 82.84M parameters and a competitive speed of 10.56 FPS, it compromises on tracking performance. 

In contrast, CDRMT shows only a moderate increase in parameters while achieving substantial performance improvements, demonstrating an optimal balance between model capacity and computational efficiency. The competitive inference speed (10.11 FPS) further  validates our framework's practicality in real-world applications, where both accuracy and efficiency are crucial. These results suggest that our cognitive disentanglement approach not only enhances tracking accuracy but also maintains computational efficiency.

\begin{table}[t]
	\centering
	\renewcommand{\arraystretch}{1.4} 
	\caption{Model efficiency comparison with state-of-the-art methods.}
	\label{tab:efficiency}
	\setlength{\tabcolsep}{12.5pt} 
	\begin{tabular}{l | ccc} 
		\toprule[1.5pt] 
		Method & Params./M & FLOPs/G & FPS \\
		\midrule[1pt] 
		TransRMOT~\cite{rmot1}$_{\text{CVPR23}}$ & 124.16 & 274.25 & 10.38 \\
		iKUN~\cite{ikun}$_{\text{CVPR24}}$ & 94.93 & 496.98 & 8.46 \\
		TempRMOT~\cite{rmot2}$_{\text{arxiv24}}$ & 130.48 & 588.70 & 8.79 \\
		MGLT-CO-MOT~\cite{rmot-tim}$_{\text{TIM25}}$ & 82.84 & 338.17 & 10.56 \\
		CDRMT (Ours) & 126.00 & 287.03 & 10.11 \\
		\bottomrule[1.5pt] 
	\end{tabular}
\end{table}
\subsubsection{Cross Comparison on Different Datasets}
\begin{table}[t]
	\centering
	\renewcommand{\arraystretch}{1.2}
	\setlength{\extrarowheight}{1.1pt}
	\caption{Cross performance comparison on Refer-KITTI and Refer-KITTI-V2 datasets. TempRMOT$^{\ddagger}$ results reproduced using official code and weights.}
	\label{tab:cross-comparison}
	\begin{tabular}{l|ccc|ccc}
		\toprule[1.5pt] 
		\multirow{2}{*}{\centering\vphantom{Hg} Methods} 
		& \multicolumn{3}{c|}{Refer-KITTI} 
		& \multicolumn{3}{c}{Refer-KITTI-V2}  \\
		\cline{2-7}
		& $\vcenter{\hbox{HOTA$\uparrow$}}$ 
		& $\vcenter{\hbox{DetA$\uparrow$}}$ 
		& $\vcenter{\hbox{AssA$\uparrow$}}$ 
		& $\vcenter{\hbox{HOTA$\uparrow$}}$ 
		& $\vcenter{\hbox{DetA$\uparrow$}}$ 
		& $\vcenter{\hbox{AssA$\uparrow$}}$ \\
		\midrule[1pt] 
		TransRMOT~\cite{rmot1} & 46.56 & 37.97 & 57.33 & 31.00 & 19.40 & 49.68 \\
		CDRMT (ours) & \textbf{49.35} & \textbf{40.34} & \textbf{60.56} & \textbf{31.99} & \textbf{20.37} & \textbf{50.35} \\
		\midrule[0.5pt] 
		TempRMOT$^{\ddagger}$~\cite{rmot2} & 50.68 & 38.75& 66.45 & 34.60 & 22.91 & 52.39 \\
		CDRMT$^{\dagger}$ (ours) & \textbf{52.04} & \textbf{40.14} & \textbf{67.59} & \textbf{35.23} & \textbf{23.55} & \textbf{52.82} \\
		\toprule[1.5pt]
	\end{tabular}
\end{table}
\begin{figure*}[!t]
	\centering
	\includegraphics[width=\linewidth]{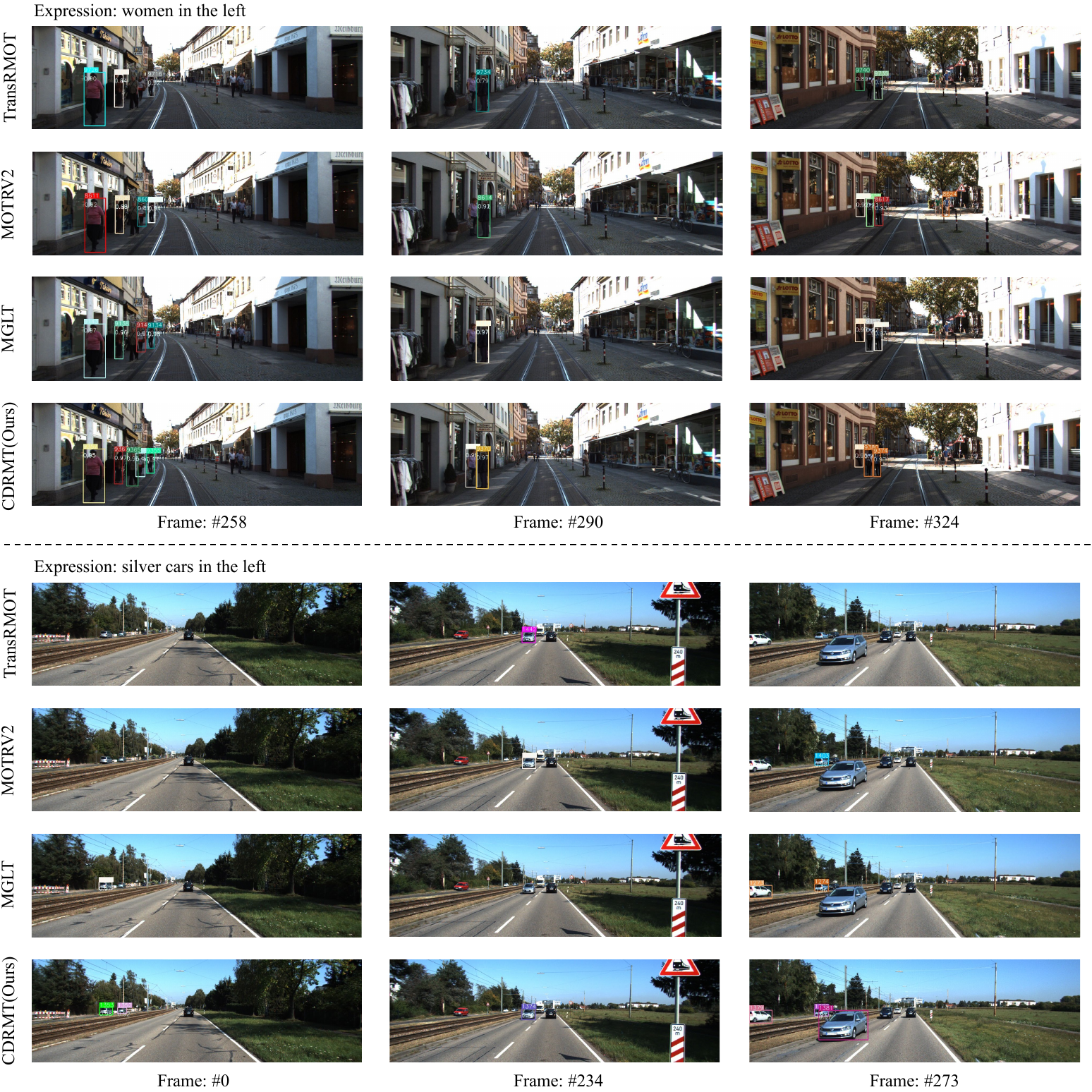}
	\caption{
		Qualitative comparison of referring tracking results with state-of-the-art methods on challenging sequences. For each sequence, we show detection and tracking results across consecutive frames, where different colors denote distinct tracking identities. Our CDRMT consistently outperforms existing methods in both accuracy and tracking robustness.  }
	\label{fig:vis1}
	\vspace{-0.5cm}
\end{figure*}
To comprehensively validate the effectiveness of our cognitive disentanglement method, we conducted extensive cross-method experiments by integrating CDRMT into two representative RMOT methods. As shown in Table~\ref{tab:cross-comparison}, after incorporating our framework into TransRMOT~\cite{rmot1}, CDRMT demonstrates substantial improvements with relative gains of 6.0\% HOTA, 6.2\% DetA, and 5.6\% AssA on Refer-KITTI, highlighting the benefits of disentangled semantic processing. More significantly, when enhancing TempRMOT$^{\ddagger}$~\cite{rmot2} (denoted as CDRMT$^\dagger$), our method still yields consistent performance boosts, albeit with comparatively smaller relative gains of 2.7\% HOTA (from 50.68 to 52.04), 3.6\% DetA (from 38.75 to 40.14), and 1.7\% AssA (from 66.45 to 67.59). This performance differential between TransRMOT and TempRMOT integration stems from several factors. TempRMOT begins from a substantially higher baseline performance, typically leading to diminishing returns as models approach higher performance levels. Additionally, TempRMOT's sophisticated temporal query memory mechanism likely captures some motion information targeted by our spatial-motion pathway, creating partial representational overlap.
Our primary goal in the transfer study was to demonstrate the general applicability of our approach rather than to maximize performance gains through extensive architectural tuning. Nevertheless, the improvements in both detection accuracy (DetA) and association quality (AssA) demonstrate that our cognitive-inspired design effectively enhances both spatial and temporal feature representations. Furthermore, these performance gains are consistently maintained on the more challenging Refer-KITTI-V2 dataset, where CDRMT$^\dagger$ achieves improvements of 1.8\% HOTA, 2.8\% DetA, and 0.8\% AssA. These substantial and consistent improvements across different baseline methods and datasets validate that our cognitive-inspired disentangled processing of static and dynamic information can effectively enhance existing RMOT frameworks, leading to more robust and accurate tracking performance in complex scenarios.

\subsection{Qualitative Results}
To validate the effectiveness of our proposed cognitive disentanglement framework, we conduct qualitative comparisons with state-of-the-art methods including TransRMOT~\cite{rmot1}, MOTRv2~\cite{motrv2}, and MGLT\cite{rmot-tim} on challenging scenarios from the Refer-KITTI dataset, as shown in Figure~\ref{fig:vis1}.
\begin{figure*}[!t]
	\centering
	\includegraphics[width=\linewidth]{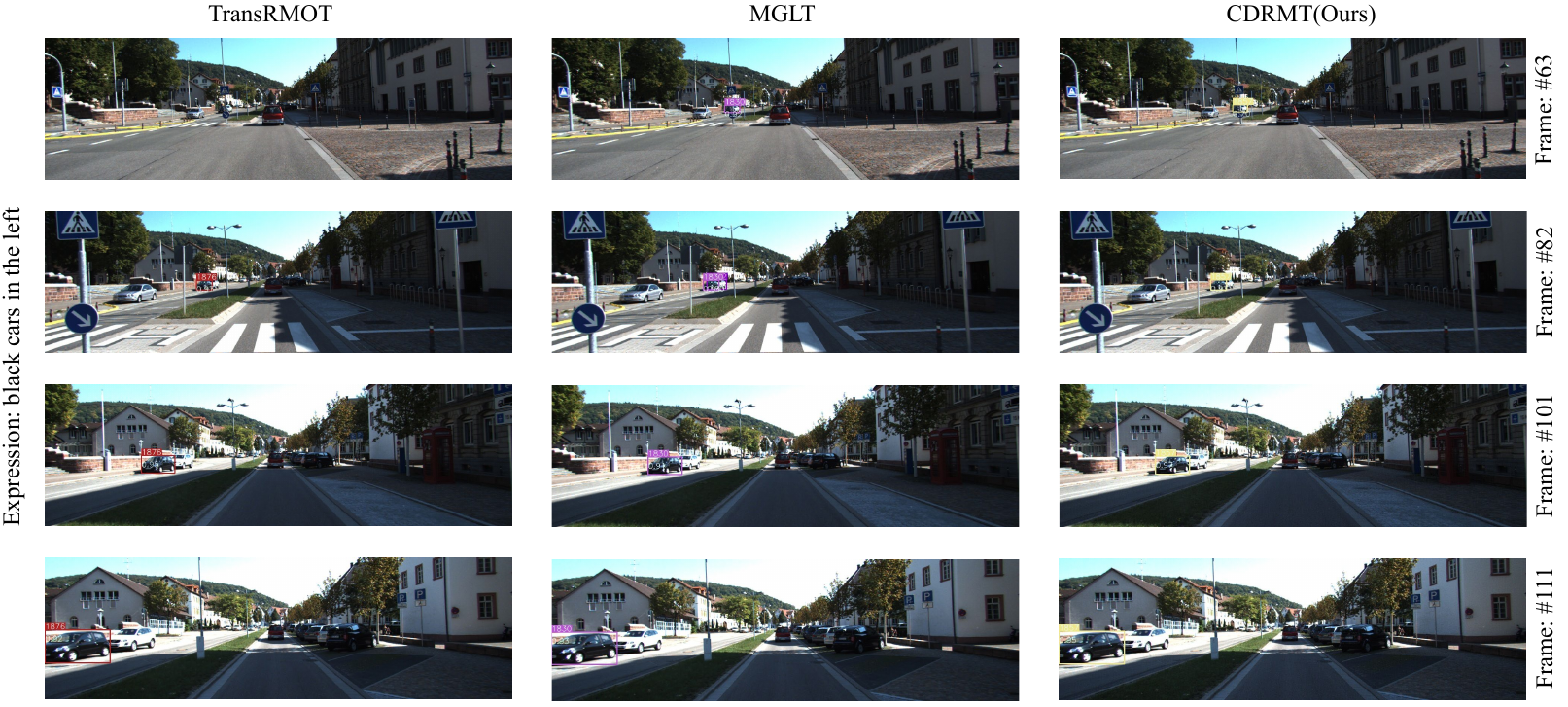}
	\caption{
		More qualitative results of the proposed CDRMT on the Refer-KITTI~\cite{rmot1} dataset. The strong alignment between natural language descriptions and tracked objects validates the capability of CDRMT in joint visual-linguistic reasoning.}
	\label{fig:vis2}
	\vspace{-0.5cm}
\end{figure*}
\begin{figure*}[!t]
	\centering
	\includegraphics[width=\linewidth]{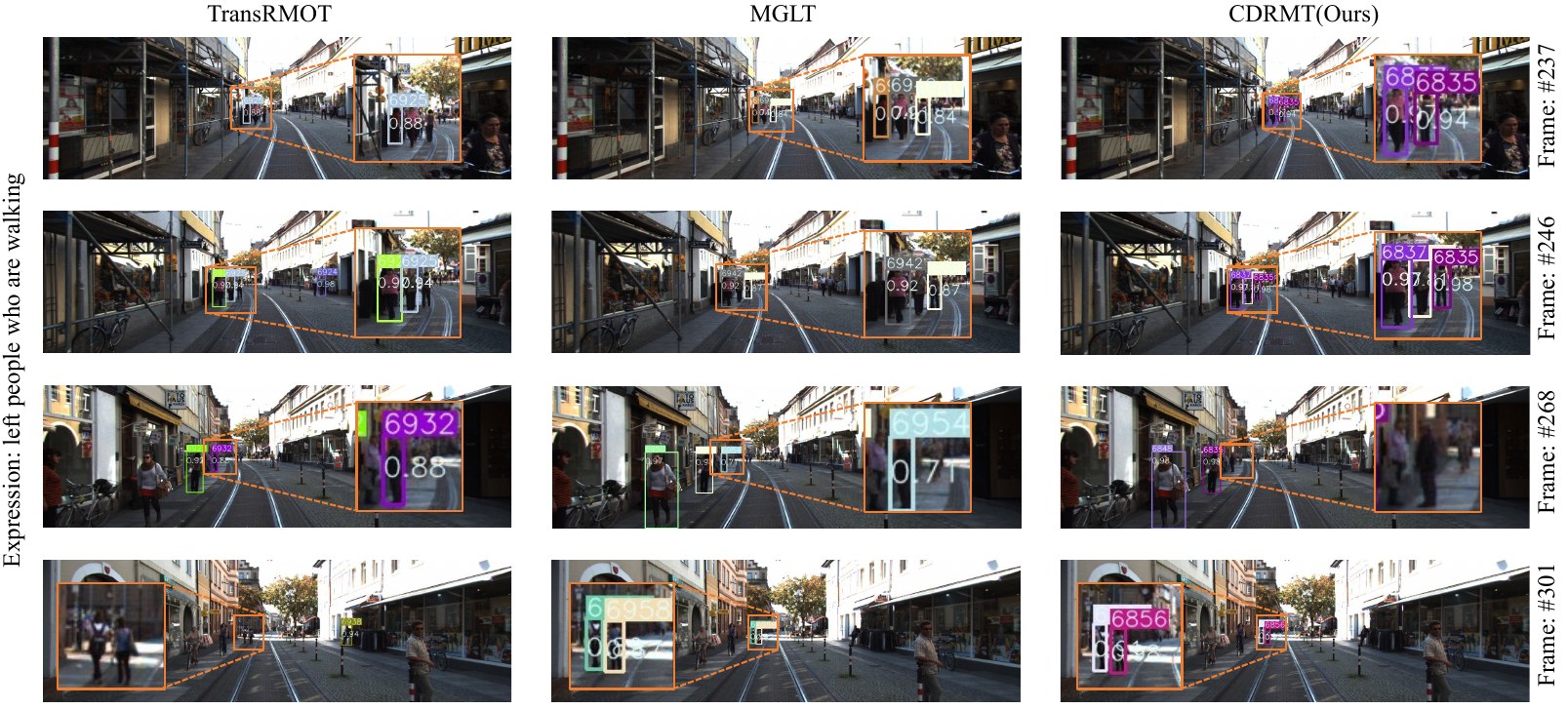}
	\caption{
		Qualitative comparison on challenging scenes with the complex expression "left people who are walking". Our CDRMT demonstrates superior performance in scenarios requiring joint understanding of static object attributes and spatial motion information.}
	\label{fig:vis3}
	\vspace{-0.5cm}
\end{figure*}

\begin{figure*}[!t]
	\centering
	\includegraphics[width=\linewidth]{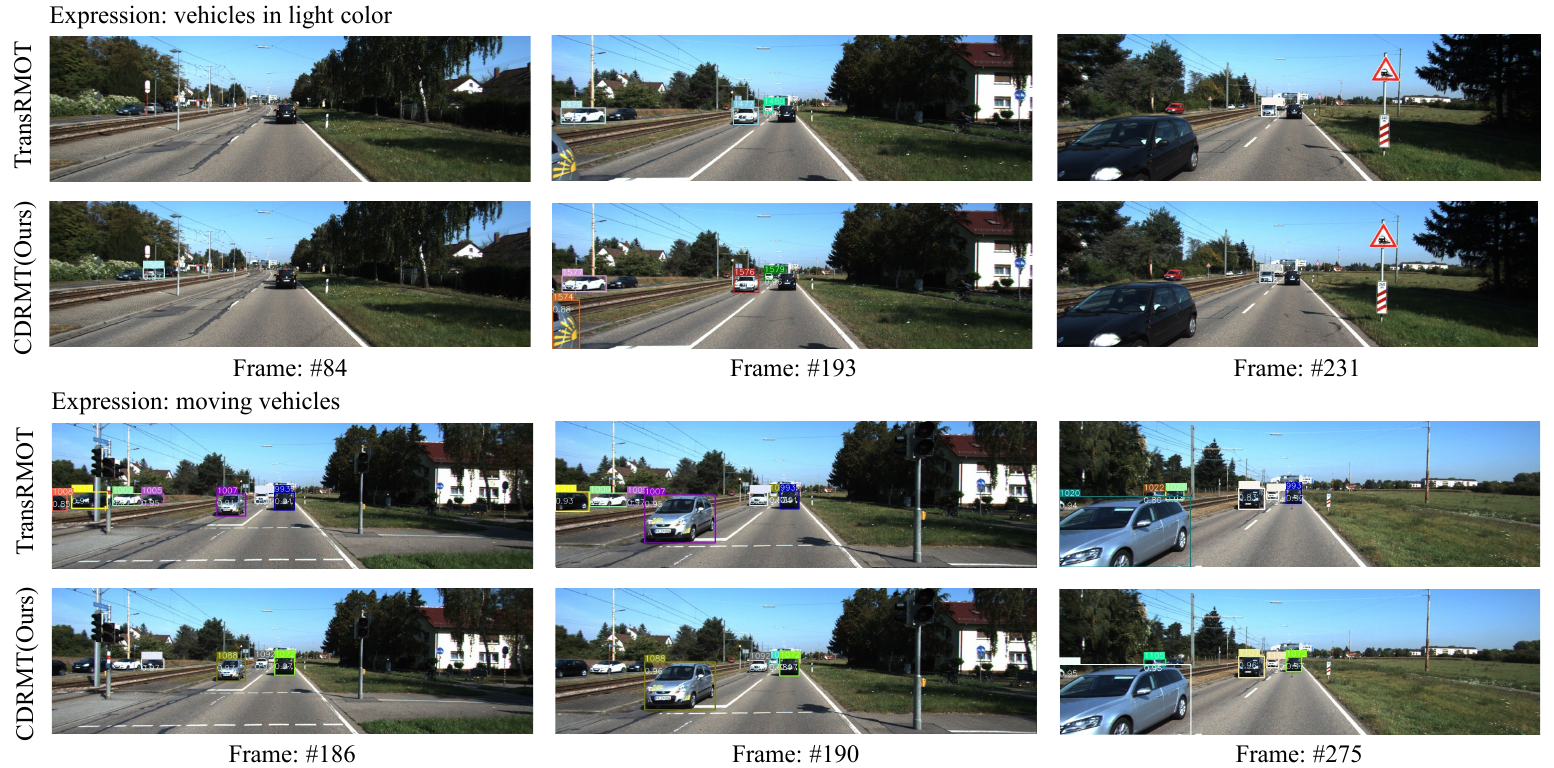}
	\caption{
		Qualitative comparison of referring tracking results with baseline methods on expressions without explicit spatial information.}
	\label{fig:vis6}
	\vspace{-0.5cm}
\end{figure*}

As illustrated in the upper example of Figure~\ref{fig:vis1}, we observe distinct performance variations in response to the referring expression "women in the left". TransRMOT exhibits substantial false negatives (FN) across all frames. Specifically, in frame $\#258$, it misses two female pedestrians in the object regions, and similar detection omissions persist in frame $\#290$. While the method detects two objects in frame $\#324$, it overlooks a crucial object near the left wall. Moreover, the erroneous attention of MOTRv2 to a male pedestrian on the right indicates insufficient or potentially misaligned integration of linguistic and visual features. Although both MOTRv2 and MGLT demonstrate comparable detection coverage, they still manifest detection insufficiencies across sequential frames. Notably, MOTRv2 exhibits substantially lower confidence scores compared to MGLT, which can be attributed to MGLT's more sophisticated language-vision feature fusion mechanism. This performance gap underscores the importance of fine-grained multimodal feature integration in referring tracking tasks.

Our proposed CDRMT significantly outperforms these baseline methods, achieving precise object localization with high confidence scores, particularly evident in frames $\#258$ and $\#324$ where it successfully captures all referred objects. While a female cyclist on the left in frame $\#290$ remains undetected, this can be primarily attributed to challenging environmental conditions, specifically suboptimal lighting and limited resolution, representing an exceptionally difficult case. This comprehensive performance analysis demonstrates CDRMT's superior capability in maintaining robust tracking under varying conditions while effectively leveraging linguistic cues for accurate object identification. This superior performance stems from our progressive semantic injection mechanism, which enables better integration of appearance features and spatial relationships. 

Moreover, Figure~\ref{fig:vis2} presents additional qualitative results comparing our CDRMT with state-of-the-art methods on the Refer-KITTI dataset, given the expression "black cars in the left". The visualization reveals distinct performance characteristics across different methods. TransRMOT exhibits significant detection instability, particularly manifesting false negatives in the initial frame and consistently low confidence scores throughout the sequence. While MGLT demonstrates improved object localization compared to TransRMOT, it suffers from imprecise bounding box regression in the first frame, where it incorrectly includes a white vehicle in the detection region. Furthermore, MGLT produces a false positive detection of a white vehicle in the second frame, indicating insufficient discrimination of color attributes specified in the referring expression. In contrast, our proposed CDRMT achieves superior performance in both object localization and attribute discrimination. The method demonstrates robust detection capabilities by accurately identifying black vehicles while maintaining precise bounding box boundaries. This enhanced performance can be attributed to our cognitive disentanglement framework, which effectively processes both static attributes ("black") and spatial information ("in the left") through separate pathways. 

Furthermore, we evaluated the performance of various methods in challenging scenes with complex expressions, as shown in Figure~\ref{fig:vis3}.
The referring expression "left people who are walking" presents a significant challenge as it requires models to simultaneously understand object categories ("people"), spatial positioning ("left"), and motion states ("walking"). In frame \#237, TransRMOT overlooks the rightmost pedestrian among the left group, failing to comprehend the motion component of the description. Meanwhile, MGLT incorrectly identifies stationary pedestrians on the left, demonstrating an inability to properly distinguish between static and dynamic objects. Similar issues persist in frame \#268, where both TransRMOT and MGLT erroneously detect stationary pedestrians, indicating insufficient perception of motion information within the language description. In contrast, CDRMT accurately identifies only those pedestrians exhibiting walking behavior, effectively filtering out stationary individuals despite their spatial proximity. In frames \#246 and \#301, TransRMOT consistently misses walking pedestrians while incorrectly detecting pedestrians on the right side. While MGLT shows improved detection capabilities compared to TransRMOT, its performance still falls short of our CDRMT. These results validate the effectiveness of our cognitive disentanglement approach in handling complex language descriptions that combine both static object attributes and spatial motion information. 

As illustrated in Figure~\ref{fig:vis6}, we evaluate our model against baselines on expressions lacking explicit spatial localization terms. For the expression "vehicles in light color," which primarily tests static attribute recognition capability, TransRMOT exhibits significant detection instability. It fails to identify the target objects in frames \#84 and \#193, and even when detection occurs in frame \#231, it shows substantially lower confidence (0.89) compared to our method (0.98). In contrast, CDRMT consistently identifies all light-colored vehicles across all frames with high confidence scores. It demonstrates the effectiveness of our enhanced static attribute processing through the "what" pathway, even in the absence of spatial indicators.   The expression "moving vehicles," which requires understanding of motion dynamics without explicit positional cues, the limitations of baseline methods become more pronounced. TransRMOT incorrectly identifies stationary vehicles as targets across frames \#186, \#190, and \#275, failing to distinguish motion attributes. Our CDRMT framework, however, correctly recognizes that motion terms like "moving" should be processed through the "where" pathway despite lacking traditional spatial prepositions. By effectively decoupling this motion information, CDRMT successfully identifies only the vehicles in motion while filtering out stationary vehicles that don't match the specified criteria. This demonstrates our model's superior capability in handling complex semantic relationships that combine both static object recognition and dynamic state understanding.
\begin{table}[t]
        \centering
        \caption{Ablation studies of different components in CDRMT. In the table, `w.' indicates `with' (component used) and `w/o.' indicates `without' (component not used). The best results are in \textbf{bold}.}
        \label{tab:ablation}
        \renewcommand{\arraystretch}{1.3}
        \setlength{\tabcolsep}{6pt}
        \begin{tabular}{c|ccc|cccccccc}
            \toprule[1.5pt]
            & \multicolumn{3}{c|}{\textbf{Components}} & \multicolumn{8}{c}{\textbf{Metrics}} \\
            \cline{2-12}
            \raisebox{-0.5ex}{} & BIF & PSDQL & SCC & HOTA$\uparrow$ & DetA$\uparrow$ & AssA$\uparrow$ & DetRe$\uparrow$ & DetPr$\uparrow$ & AssRe$\uparrow$ & AssPr$\uparrow$ & LocA$\uparrow$\\
            \midrule[1pt]
            \ding{172} & \xmark & \xmark & \xmark & 46.56 & 37.97 & 57.33 & 49.69 &\textbf{60.10} & 60.02 & 89.67 & 90.33\\
            \ding{173} & \cmark & \xmark & \xmark & 46.96 & 38.65 & 57.11 &55.75 & 54.64 & 61.35 & 88.73 & 90.43\\
            \midrule[1pt]
            \ding{174} & \xmark & \cmark & \xmark & 48.36 & 39.97 & 58.69 & 54.86 & 58.07 & 62.79 & 89.68 & 90.32\\
            \ding{175} & \xmark & w/o. ${Q}'_{t}$ & \xmark & 48.02 & 39.45 & 58.62 & 55.46 & 56.41 & 63.09 & 88.77 & 90.52\\
            \ding{176} & \xmark & w/o. $\hat{Q}'_{t}$ & \xmark & 47.89 & 39.63 & 58.04 & 54.66 & 57.64 & 61.94 & 89.77 & 90.60\\
            \midrule[1pt]
            \ding{177} & \cmark & \cmark & \xmark & 48.67 & 40.05 & 59.31 & \textbf{55.87} & 57.89& 63.13 & 88.57 & 90.51\\
            \ding{178} & \cmark & \xmark & \cmark & 47.71 & 39.13 & 58.34 & 53.97 & 55.43& 62.61 & 88.75 & 90.23\\
            \ding{179} & \xmark & \cmark & \cmark & 49.06 & 40.11 & 60.13 & 54.97 & 58.72& 62.95 & 89.24 & 90.57\\
            \ding{180} & \cmark & \cmark & \cmark &  \textbf{49.35} & \textbf{40.34} & \textbf{60.56} & 54.54 & 59.30 & \textbf{64.70} & \textbf{89.80} & \textbf{90.61} \\
            \toprule[1.5pt]
        \end{tabular}
\end{table}
\subsection{Ablation Studies}
\subsubsection{Evaluation of Components}
To thoroughly investigate the effectiveness of each proposed component and validate our design choices, we conduct comprehensive ablation studies on the Refer-KITTI. The detailed experimental results are presented in Table~\ref{tab:ablation}.
The baseline architecture (row \ding{172}), as our comparison, we first examine the effect of integrating the Bidirectional Interactive Fusion (BIF) module (row \ding{173}). This integration yields quantifiable improvements in holistic tracking accuracy (HOTA increases by 0.9\%) and detection performance (DetA improves by 1.8\%). These enhancements empirically validate our hypothesis that structured bidirectional cross-modal interactions prior to the encoder stage facilitate more effective modality-specific feature refinement while preserving representational integrity.
Further incorporating the Progressive Semantic-Decoupled Query Learning (PSDQL) module (row \ding{174}) is analogous to the processing of the dorsal stream and ventral stream in human visual system. We observe consistent and substantial gains across all evaluation metrics (HOTA: 48.36, DetA: 39.97, AssA: 58.69). 
To rigorously validate the necessity of our progressive design, we conduct controlled experiments by ablating each stream independently. When removing the static object query ${Q}'_{t}$ (row \ding{175}), which is responsible for encoding appearance and attribute information, the HOTA metric drops to 48.02. This degradation is particularly evident in detection accuracy (DetA decreases by 1.3\%), indicating that the model's ability to precisely locate objects based on their visual attributes is compromised. Similarly, ablating the spatial motion information query $\hat{Q}'_{t}$ (row \ding{176}) leads to a performance decline (HOTA: 47.89), with a more pronounced impact on association accuracy (AssA decreases by 1.1\%). This suggests that without explicit motion state modeling, the model struggles to maintain consistent object identities across frames, especially in challenging scenarios with complex motion patterns or object interactions.
\begin{figure*}[!t]
	\centering
	\includegraphics[width=\linewidth]{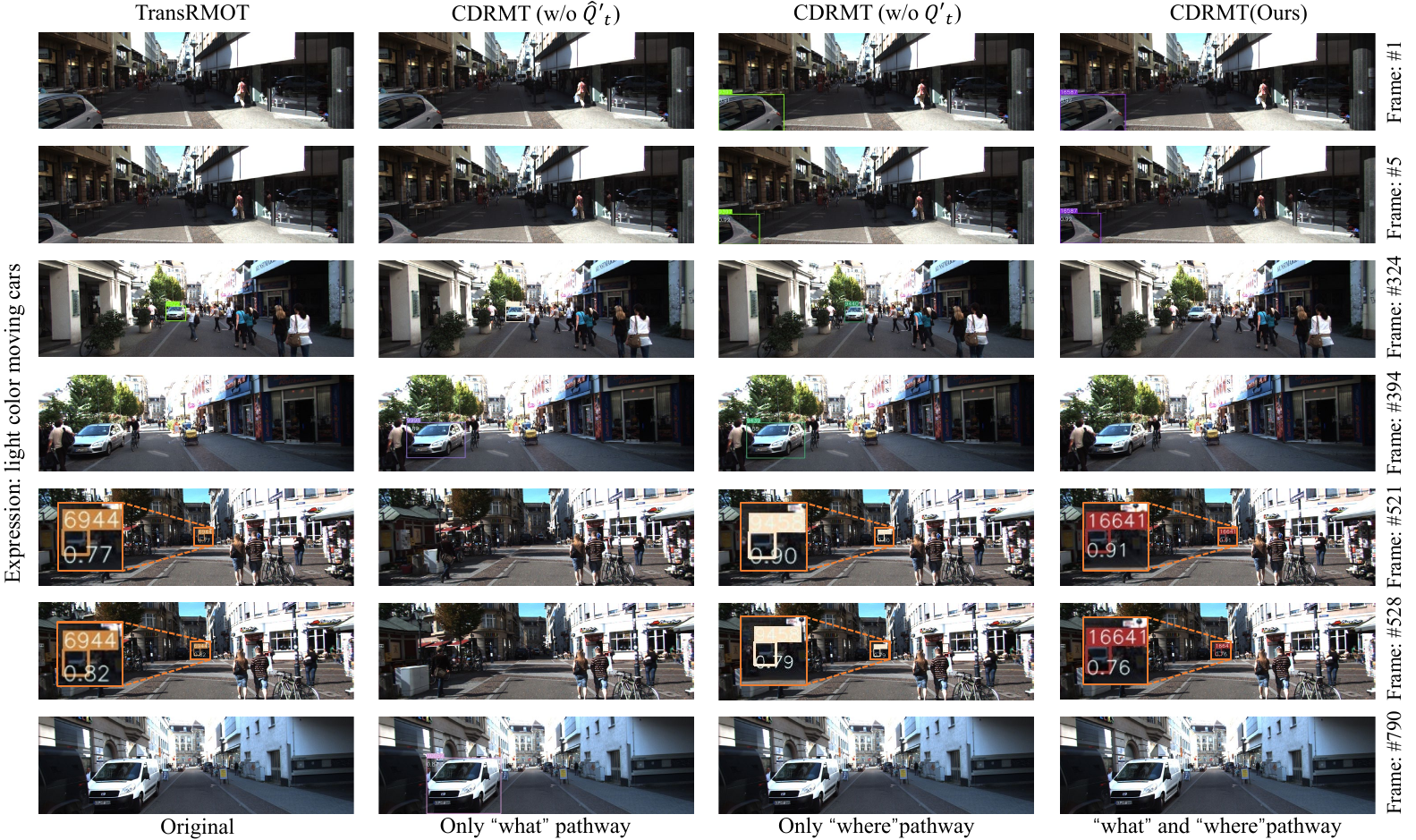}
	\caption{
		Qualitative comparison of tracking results across different model variants on challenging examples. From left to right: TransRMOT (baseline), CDRMT without spatial motion query processing ($\hat{Q}'_{t}$), CDRMT without static object query processing (${Q}'_{t}$), and our  CDRMT model. }
	\label{fig:vis4}
	\vspace{-0.5cm}
\end{figure*}

The complementary nature of these performance drops aligns well with neuroscience findings about the ventral (``what'') and dorsal (``where'') streams in human visual processing. Just as the human visual system relies on both pathways for comprehensive scene understanding, as shown in Figure~\ref{fig:vis4}, our two query streams each make unique and substantial contributions to overall tracking performance. The static object query stream (ventral pathway) provides crucial appearance-based discrimination capabilities. In Figure~\ref{fig:vis4}, CDRMT (w/o $\hat{Q}'_{t}$) demonstrates enhanced ability to recognize static feature objects like "light color cars" compared to the baseline TransRMOT, due to the injection of "what" pathway information. However, without guidance from dynamic information, this variant fails to effectively track moving objects in frames \#1, \#5, and frames \#521, \#528. Additionally, overreliance on static information leads to false detection of non-object objects in frames \#324, \#394, and \#790. In contrast, CDRMT (w/o ${Q}'_{t}$) shows superior ability to identify dynamic objects like "moving cars" compared to the baseline, successfully tracking object movement between frames due to the injection of "where" pathway information. While it incorrectly identifies some static "light color cars" in frames \#324 and \#394 , it successfully corrects this in frame \#790. Our complete CDRMT method employs a progressive information injection strategy (coordinating "what" and "where" pathways), accurately identifying all object objects while effectively filtering non-object objects, maintaining stable tracking performance across all test frames. This result strongly demonstrates that our proposed progressive learning mechanism is essential for handling the complex interactions between static visual attributes and spatial motion information in RMOT tasks, validating the effectiveness of our cognitive disentanglement approach inspired by the human visual system in complex visual tracking tasks.

We further investigate the interaction effects between multiple components to evaluate their complementarity. The joint implementation of BIF and PSDQL (row \ding{177}) yields performance superior to either component in isolation (HOTA: 48.67, DetA: 40.05, AssA: 59.31). 
The combination of BIF and Structural Consensus Constraint (SCC) (row \ding{178}) demonstrates moderate improvement over the BIF-only configuration, while the integration of PSDQL with SCC (row \ding{179}) produces substantial performance gains (HOTA: 49.06). The experimental results demonstrate that SCC not only helps maintain robust tracking performance but also enhances the model’s interpretability by enforcing semantic alignment between visual tracking results and language descriptions.

Our complete model (CDRMT), integrating all the proposed components, achieves the optimal performance across the majority of evaluation metrics. These results comprehensively validate the effectiveness of our cognitive science-inspired approach to RMOT tasks and demonstrate the synergistic benefits of combining multiple specialized processing streams with semantic consistency constraints.

\subsubsection{Number of Detection Queries}
Table \ref{tab:queries} investigates the impact of varying the number of detection queries on tracking performance. We conduct experiments with different numbers of queries (60, 100, 200, and 300) while keeping other components unchanged. The results demonstrate that the tracking performance consistently improves as the number of queries increases, indicating the importance of sufficient query capacity in handling RMOT tasks.

\begin{table}[t]
	\centering
	\caption{Impact of different number of queries.}
	\label{tab:queries}
	\begin{tabular}{c|cccccccc}
		\toprule[1.5pt]
		Queries Numbers & HOTA$\uparrow$ & DetA$\uparrow$ & AssA$\uparrow$ & DetRe$\uparrow$ & DetPr$\uparrow$ & AssRe$\uparrow$ & AssPr$\uparrow$ & LocA$\uparrow$ \\
		\midrule[1pt]
		60 & 39.03 & 29.38 & 51.85 & 41.17 & 46.24 & 56.77&80.15 & 79.85 \\
		100 & 42.58 & 34.57 & 52.44 & 49.83 & 50.98 & 59.21 &86.29 & 88.71\\
		200 & 48.08 & 39.45 & 58.61 & 55.45 & 56.40 & 63.08 &88.63 & 90.05\\
		300 & \textbf{49.35} & \textbf{40.34} & \textbf{60.56} & \textbf{54.54} & \textbf{59.30} & \textbf{64.70} & \textbf{89.80} & \textbf{90.61} \\
		\bottomrule[1.5pt]
	\end{tabular}
\end{table}
\begin{table}[t]
		\centering
		\caption{Impact of different referring thresholds $\delta$.}
		\label{tab:thresholds}
		\begin{tabular}{c|cccccccc}
			\toprule[1.5pt]
			$\delta$ & HOTA$\uparrow$ & DetA$\uparrow$ & AssA$\uparrow$&DetRe$\uparrow$ & DetPr$\uparrow$ & AssRe$\uparrow$ & AssPr$\uparrow$ & LocA$\uparrow$  \\
			\midrule[1pt]
			0.2 & 45.82 & 37.38 & 56.31 & 48.96& 59.81& 59.73 & 90.19 & 90.37 \\
			0.3 & 46.39 & 37.57 & 57.44& 49.83 & 58.98 & 61.21 & 90.29& \textbf{90.71}\\
			0.4 & 47.82 & 38.95 & 58.72& 52.52 & 59.01 & 62.83 & 89.64& 89.71\\
			0.5 & \textbf{49.35} & \textbf{40.34} & \textbf{60.56} & \textbf{54.54} & 59.30 & \textbf{64.70} & 89.80 & 90.61 \\
			0.6 & 48.71 & 40.15 & 59.24& 54.10 & 59.48 & 63.31 & 90.09& 90.65\\
			0.7 & 47.46 & 38.54 & 58.57& 48.89 & \textbf{62.80} & 62.41 & 89.79& 90.62\\
			0.8 & 46.06 & 37.87 & 56.22& 52.51 & 56.20 & 59.67 & \textbf{90.35}& 90.43\\
			\bottomrule[1.5pt]
		\end{tabular}
\end{table}

Specifically, when increasing queries from 60 to 300, we observe substantial improvements across all metrics: HOTA improves from 39.03 to 49.35, DetA from 29.38 to 40.34, and AssA from 51.85 to 60.56. 
The consistent improvement pattern indicates that sufficient queries are crucial for establishing robust cross-modal correspondences between language descriptions and visual objects. However, we observe that the performance gain tends to plateau when the number of queries reaches 300, suggesting that this is an optimal balance between model capacity and computational efficiency. Therefore, we adopt 300 queries as the default setting in our framework. 

\subsubsection{Weight of Referring Threshold}
Given that we have modified the model architecture with our proposed decoupled design, it becomes crucial to re-examine the impact of the referring threshold $\delta$, as different architectures may exhibit varying sensitivities to this hyperparameter. To this end, we conduct a systematic study by varying $\delta$ from 0.2 to 0.8, with results presented in Table~\ref{tab:thresholds}.
	
The experimental results reveal that our architecture demonstrates distinct response patterns to different threshold values. Specifically, when $\delta$ is set to lower values (0.2-0.4), although maintaining high AssPr (90.19) and LocA (90.37), the model shows relatively weak performance in HOTA (45.82) and DetA (37.38). This suggests that our enhanced feature representation requires a more stringent threshold to filter out potential false matches. As $\delta$ increases to 0.5, we observe optimal performance across multiple metrics, with HOTA reaching 49.35, DetA improving to 40.34, and AssA achieving 60.56. However, further increasing $\delta$ to 0.8 leads to performance degradation, with HOTA dropping to 46.06. This comprehensive analysis helps us identify $\delta=0.5$ as the optimal threshold for our architecture, ensuring robust performance in referring tracking tasks.
\begin{table}[t]
		\centering
		\caption{Impact of different components of  Query Semantic Injection (QSI) module.}
		\label{tab:ablation_QSI}
		\begin{tabular}{c|ccc}
			\toprule[1.5pt]
			QSI & HOTA$\uparrow$ & DetA$\uparrow$ & AssA$\uparrow$  \\
			\midrule[1pt]
			- & 47.16 & 38.82 & 57.47  \\
			\midrule
			$\hat{\mathbf{f}}_{so}$ \& $\hat{\mathbf{f}}_{sm}$ & 47.31 & 39.02 & 57.82 \\
			$\hat{\mathbf{f}}_{sm}$ $\rightarrow$ $\hat{\mathbf{f}}_{so}$ & 47.63 & 39.37 & 58.17 \\
			$\hat{\mathbf{f}}_{so}$ $\rightarrow$ $\hat{\mathbf{f}}_{sm}$ & 48.06 & 39.77 & 58.22 \\
			$Q^{Det}_{inject}$ & 48.86 & 40.22 & 59.42 \\
			$Q^{Tra}_{inject}$ & 48.53 & 39.54 & 60.17 \\
			$Q^{Det}_{inject}$ \& $Q^{Tra}_{inject}$ & \textbf{49.35} & \textbf{40.34} & \textbf{60.56} \\
			\bottomrule[1.5pt]
		\end{tabular}
\end{table}

\begin{table}[t]
	\centering
	\caption{Ablation study on Structural Consistency Constraint (SCC) module.}
	\label{tab:ablation_scc}
	\begin{tabular}{c|ccc}
		\toprule[1.5pt]
		SCC & HOTA$\uparrow$ & DetA$\uparrow$ & AssA$\uparrow$ \\
		\midrule
		- & 47.67 & 39.37 & 58.14 \\
		\midrule
		Point-wise Consistency (PwC) & 47.75 & 39.46 & 58.39 \\
		Distance Consistency (DC) & 48.41 & 39.83 & 59.29 \\
		Angle Consistency (AC) & 48.19 & 39.65 & 58.82 \\
		DC+AC ($\lambda_{angle}=0.2$) & 48.83 & 40.07 & 59.78 \\
		DC+AC ($\lambda_{angle}=0.4$) & \textbf{49.35} & \textbf{40.34} & \textbf{60.56} \\
		DC+AC ($\lambda_{angle}=0.6$) & 49.07 & 40.29 & 59.97 \\
		\toprule[1.5pt]
	\end{tabular}
\end{table}
\subsubsection{Different Components of QSI Module}
Table~\ref{tab:ablation_QSI} presents a comprehensive ablation study of our Query Semantic Injection (QSI) module. Without QSI, the model achieves performance (47.16 HOTA, 38.82 DetA, 57.47 AssA), indicating the limitations of conventional feature fusion approaches. When simultaneously injecting both static object and spatial motion information ($\hat{\mathbf{f}}_{so}$ \& $\hat{\mathbf{f}}_{sm}$), we observe initial improvements in all metrics (47.31 HOTA, 39.02 DetA, 57.82 AssA). However, this parallel injection strategy, while beneficial, fails to fully exploit the hierarchical nature of semantic information.
In exploring semantic injection ordering, we find that injecting spatial motion information before static object attributes ($\hat{\mathbf{f}}_{sm}$ $\rightarrow$ $\hat{\mathbf{f}}_{so}$) leads to suboptimal performance (47.63 HOTA), as the model struggles to establish stable object representations without first understanding basic object characteristics. Conversely, following the cognitive process of human vision by first injecting static object features followed by motion information ($\hat{\mathbf{f}}_{so}$ $\rightarrow$ $\hat{\mathbf{f}}_{sm}$) yields superior results (48.06 HOTA), demonstrating the importance of progressive semantic understanding from basic to complex features.
The impact of query-specific injection reveals interesting patterns. Detection query injection ($Q^{Det}_{inject}$) significantly enhances localization accuracy (40.22 DetA) by providing explicit semantic guidance for object detection. Meanwhile, track query injection ($Q^{Tra}_{inject}$) exhibits stronger temporal modeling capabilities, as evidenced by improved association scores (60.17 AssA). The synergistic combination of both injection strategies ($Q^{Det}_{inject}$ \& $Q^{Tra}_{inject}$) achieves optimal performance across all metrics (49.35 HOTA, 40.34 DetA, 60.56 AssA), validating our hypothesis that comprehensive semantic understanding requires fine-grained semantic guidance.

\subsubsection{Different Components of SCC Module}
Table~\ref{tab:ablation_scc} presents the impact of different Structural Consistency Constraint (SCC) configurations on model performance. The baseline model (``-'') without any consistency constraint performs worst across all metrics. The traditional Point-wise Consistency (PwC), which only operates at the single sample level by directly comparing the original text embedding $\mathbf{e}$ with its corresponding reconstructed text embedding $\mathbf{e}'$, shows limited improvement (HOTA increases by only 0.16\%).
In contrast, our proposed structural consistency approach includes two key components.  The Distance Consistency (DC) constraint leverages relationships between three different text embeddings (static object embedding $\mathbf{e}_{so}$, spatial motion embedding $\mathbf{e}_{sm}$, and global sentence representation $\mathbf{e}_g$) and their reconstructed counterparts. By maintaining the relative distance structure between these embedding pairs rather than direct one-to-one mapping, it achieves significant improvement (HOTA reaches 48.41). Similarly, the Angle Consistency (AC) constraint considers angular relationships formed by triplets of embeddings, capturing higher-order semantic structures and achieving a HOTA of 48.19.
The most significant performance gains come from combining both constraints into a complete Structural Consistency Constraint (SCC). When $\lambda_{angle}=0.4$ , the model achieves optimal performance across all metrics. This demonstrates the strong complementarity between the two constraints: distance consistency preserves semantic similarities, while angle consistency captures more complex structural relationships between embeddings.
\begin{figure*}[!t]
	\centering
	\includegraphics[width=\linewidth]{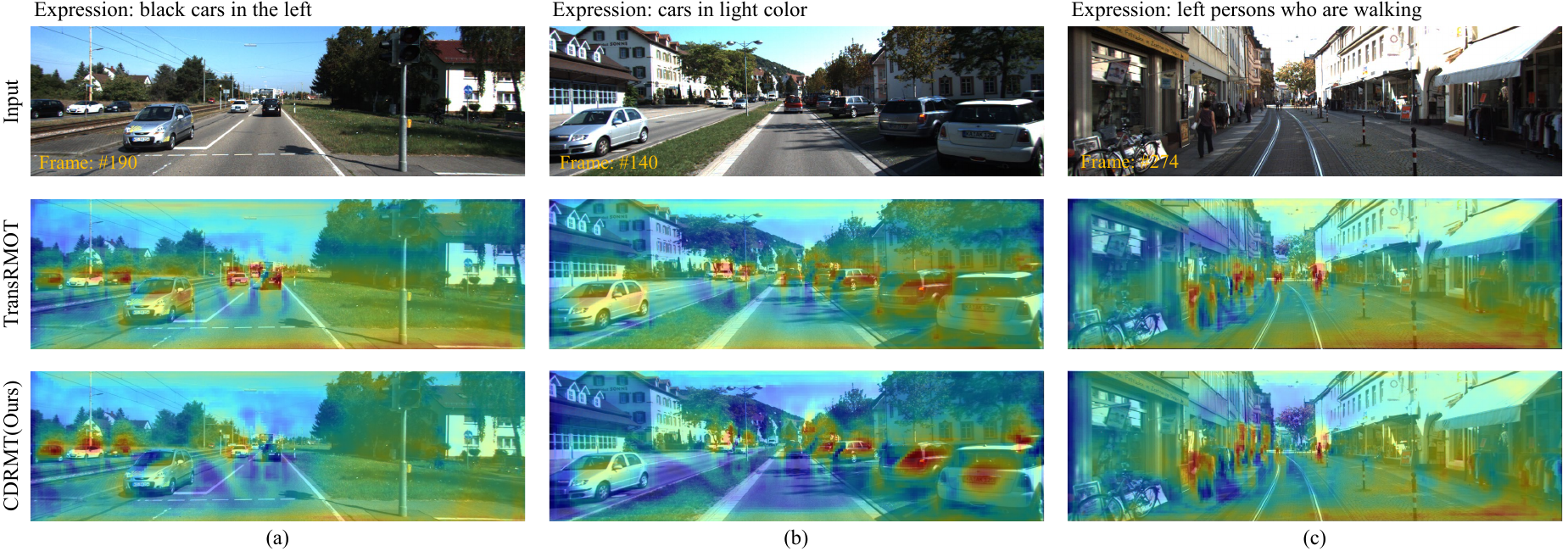}
    \vspace{-0.5cm}
	\caption{
		Visualization of heatmap from our CDRMT framework compared with baseline TransRMOT. The color intensity represents attention activation strength, with warmer colors (red and yellow) indicating regions of higher feature activation and cooler colors (blue) indicating regions of lower activation.}
	\label{fig:vis}
\end{figure*}
\section{Visualization}
\label{sec:vis}
To illustrate the effectiveness of our proposed CDRMT framework, we visualize the attention heatmaps from the final encoder layer in Figure~\ref{fig:vis}. The integration of our method results in significantly more focused attention on regions that correspond to the referring text. As shown in Figure~\ref{fig:vis}(a) with the expression "black cars in the left," while both models show attention on a white car positioned on the left, our model demonstrates heightened attention specifically towards the upper part of the vehicle where the roof is black, indicating a more nuanced understanding of the object's attributes. This more targeted attention distribution is semantically aligned with the "black cars" description, even when the car body is predominantly white. Similarly, in Figure~\ref{fig:vis}(b) with "cars in light color," the baseline model exhibits a more uniform attention distribution across the scene, whereas our approach shows concentrated attention specifically on vehicles with lighter color tones, demonstrating improved semantic discrimination. In Figure~\ref{fig:vis}(c), our model exhibits highly discriminative attention patterns that concentrate exclusively on walking pedestrians on the left side of the scene, while successfully suppressing attention on both stationary pedestrians.

\section{Discussion}
\label{sec:dis}
Overall, our Cognitive Disentanglement for Referring Multi-Object Tracking (CDRMT) framework demonstrates significant advantages in handling complex semantic information in RMOT tasks. By inspired the dual-stream processing mechanism of the human visual system, our approach effectively disentangles static object attributes from spatial motion information, enabling more precise object localization and tracking. 
\subsection{Limitations}
\begin{figure*}[!t]
	\centering
	\includegraphics[width=\linewidth]{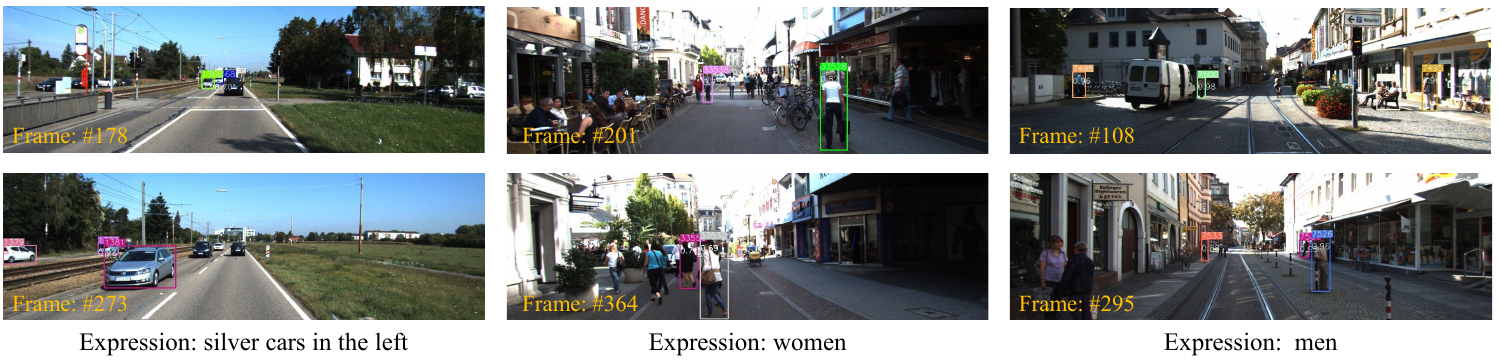}
        \vspace{-0.5cm}
	\caption{
		The failure cases of CDRMT. }
	\label{fig:failure}
	\vspace{-0.5cm}
\end{figure*}

However, we acknowledge certain limitations of our current approach, particularly in challenging environmental conditions. We have conducted further analysis of failure cases to better understand the limitations of our approach. Figure~\ref{fig:failure} illustrates several representative failure scenarios where CDRMT encounters difficulties. These include cases with extreme lighting conditions, severe occlusions, and low-resolution scenarios that challenge our cognitive disentanglement approach.
Environmental variations impact performance, particularly with color-based attributes. For instance, our model misidentifies a white car as "silver" due to strong sunlight reflections (Figure~\ref{fig:failure} \#178, \#273). Crowded scenarios with occlusions present challenges, as demonstrated by our model's failure to detect certain instances of "woman" in an outdoor cafe (Figure~\ref{fig:failure} \#201, \#364) and in a busy street scene (\#364). Low-resolution scenarios with similar objects lead to misidentifications(Figure~\ref{fig:failure} \#108, \#295), where women with shorter hairstyles appear visually similar to men from a distance. 

Our cognitive disentanglement approach effectively handles complex expressions by separating "what" and "where" pathways, but it also depends on visual feature quality. When these features become ambiguous due to resolution limitations or inherent visual similarities, even our enhanced semantic processing may fail to correctly associate with linguistic elements.
\subsection{Future Work}
Looking forward, the limitations identified in our study motivate several important directions for future research. We plan to construct a dedicated RMOT dataset focusing on challenging scenarios to better address the limitations in extreme conditions. This dataset would specifically include scenes with varying lighting conditions, partial occlusions, and complex background environments to push the boundaries of RMOT research. In addition, the current dataset expression is relatively fixed and cannot achieve dynamic and fine-grained descriptions that change over time. We hope to solve this problem in the future. Exploring multi-granularity feature integration techniques could address the challenge of similar objects in low-resolution scenarios.
The poor performance on rare motion patterns like "turning" highlights the need for specialized techniques to handle long-tailed distributions in motion expressions.  It is worth mentioning that future work could explore integrating more advanced language models with stronger contextual reasoning capabilities. This would help the system better understand implicit spatial relationships and make more human-like inferences about object relationships and behaviors.

We believe these developments will contribute to more robust and reliable RMOT systems capable of handling a broader range of real-world scenarios, ultimately bringing us closer to human-level performance in language-guided multi-object tracking tasks.
\section{Conclusion}
\label{sec:clu}
In this paper, we present a novel Cognitive Disentanglement for Referring Multi-Object Tracking (CDRMT) framework that effectively addresses the challenges of processing complex semantic information in multi-source information fusion. Drawing inspiration from the dual-stream processing mechanism in human visual system, our approach explicitly disentangles static object attributes and spatial motion information in referring expressions, enabling more precise and robust tracking results. The proposed Bidirectional Interactive Fusion module enables structured cross-modal interactions while preserving modality-specific characteristics. Additionally, the Progressive Semantic-Decoupled Query Learning mechanism hierarchically injects these two types of information into different decoder layers, while the Structural  Consistency Constraint ensures comprehensive semantic alignment between visual and linguistic modalities.  Extensive experiments on multiple benchmark datasets demonstrate that CDRMT achieves SOTA performance, which validates the effectiveness of our method.  Future research could explore the extension of our cognitive disentanglement approach to other information fusion tasks and investigate more sophisticated mechanisms for modeling the dynamic interplay between different semantic components in natural language descriptions.


\bibliographystyle{cas-model2-names}
\bibliography{cas-refs}
\end{document}